\documentclass[10pt,twocolumn,letterpaper]{article}

\usepackage[pagenumbers]{cvpr} 

\usepackage{xspace}
\usepackage{multirow}
\usepackage{colortbl}
\usepackage{graphicx}
\usepackage{booktabs}
\usepackage{siunitx}
\usepackage{duckuments}
\usepackage{pifont} 
\usepackage{makecell}
\usepackage{amsthm}
\usepackage{algorithm}
\usepackage[noend]{algpseudocode}
\usepackage{bm}
\usepackage{listings}
\usepackage{bbm}
\usepackage{xcolor}

\definecolor{LightGray}{gray}{1}

\theoremstyle{definition}

\theoremstyle{definition}

\definecolor{lightgray}{gray}{0.95}
\definecolor{midgray}{gray}{0.55}
\definecolor{steelblue}{HTML}{4D82B7}
\definecolor{davysgrey}{rgb}{0.33, 0.33, 0.33}
\definecolor{LightCyan}{rgb}{0.88,1,1}
\definecolor{LightGold}{HTML}{F3E2C5}
\definecolor{AngelRow}{HTML}{FFFDD0}
\definecolor{ao(english)}{rgb}{0.0, 0.5, 0.0}
\definecolor{lightsalmon}{rgb}{1.0, 0.63, 0.48}
\definecolor{tabgreen}{HTML}{2CA02C}
\definecolor{newgreen}{HTML}{CCE0AC}

\newcommand{\quotationmarks}[1]{``#1''}

\newcommand{\tit}[1]{\smallbreak\noindent\textbf{#1 }}

\newcommand{\tinytit}[1]{\noindent\textbf{#1}}

\usepackage{array}
\newcommand{\PreserveBackslash}[1]{\let\temp=\\#1\let\\=\temp}
\newcolumntype{C}[1]{>{\PreserveBackslash\centering}p{#1}}
\newcolumntype{R}[1]{>{\PreserveBackslash\raggedleft}p{#1}}
\newcolumntype{L}[1]{>{\PreserveBackslash\raggedright}p{#1}}
\newcolumntype{Y}{>{\centering\arraybackslash}X}

\newcommand{\deltaneg}[1]{\ensuremath{\text{\textcolor{red}{\scriptsize(#1)}}}}
\newcommand{\deltapos}[1]{\ensuremath{\text{\textcolor{green!60!black}{\scriptsize(#1)}}}}
\newcommand{\deltazero}{\ensuremath{\text{\textcolor{black}{\scriptsize(+0.0)}}}}

\definecolor{MaterialRed}{HTML}{D32F2F}        
\definecolor{MaterialBlue}{HTML}{1976D2}       
\definecolor{MaterialGreen}{HTML}{388E3C}      
\definecolor{MaterialOrange}{HTML}{F57C00}     
\definecolor{MaterialPurple}{HTML}{7B1FA2}     
\definecolor{MaterialTeal}{HTML}{00796B}       
\definecolor{MaterialIndigo}{HTML}{303F9F}     
\definecolor{MaterialBrown}{HTML}{5D4037}      

\usepackage{tcolorbox}

\definecolor{cvprblue}{rgb}{0.21,0.49,0.74}
\definecolor{darkgreen}{rgb}{0,0.6,0} 
\usepackage[pagebackref,breaklinks,colorlinks,allcolors=cvprblue]{hyperref}

\usepackage{siunitx}
\usepackage{multicol}
\usepackage{graphicx}
\usepackage{booktabs}

\usepackage{multirow}
\usepackage[dvipsnames,table]{xcolor}
\usepackage{colortbl}
\usepackage{lipsum} 

\usepackage{multirow}

\def\paperID{*****} 
\def\confName{CVPR}
\def\confYear{2026}

\title{Zero-Shot Synthetic-to-Real Handwritten Text Recognition via Task Analogies}

\author{
Carlos Garrido-Munoz\textsuperscript{1} \quad
Aniello Panariello\textsuperscript{2} \quad
Silvia Cascianelli\textsuperscript{2} \quad
Angelo Porrello\textsuperscript{2} \\
Simone Calderara\textsuperscript{2} \quad
Jorge Calvo-Zaragoza\textsuperscript{1} \quad
Rita Cucchiara\textsuperscript{2} \\
\\
\textsuperscript{1}University of Alicante, Spain \quad
\textsuperscript{2}University of Modena and Reggio Emilia, Italy
}

\graphicspath {{figures/}}
\begin{document}
\maketitle

\let\origaddcontentsline\addcontentsline
\renewcommand{\addcontentsline}[3]{}   

\begin{abstract}
Handwritten Text Recognition (HTR) models trained on synthetic handwriting often struggle to generalize to real text, and existing adaptation methods still require real samples from the target domain. In this work, we tackle the fully zero-shot synthetic-to-real generalization setting, where no real data from the target language is available. Our approach learns how model parameters change when moving from synthetic to real handwriting in one or more source languages and transfers this learned correction to new target languages. When using multiple sources, we rely on linguistic similarity to weigh their contrubition when combining them. Experiments across five languages and six architectures show consistent improvements over synthetic-only baselines and reveal that the transferred corrections benefit even languages unrelated to the sources.
\end{abstract}    
\section{Introduction}\label{sec:intro}

Handwritten Text Recognition (HTR) aims to transcribe handwritten content into machine-readable text, supporting applications from the digitization of historical archives to form automation and document retrieval. Despite notable advances in recent years~\cite{garrido2025htrsurvey}, HTR remains an open challenge due to the vast variability in human handwriting. Differences in writing style, ligatures, slant, and pressure cause models to generalize poorly across writers, scripts, and languages, leading to a persistent domain shift problem~\cite{pippi_how_2023,garrido_cvpr_2025}. Overcoming this variability typically requires large quantities of diverse, labeled handwriting data to expose models to the full spectrum of writing styles.

However, collecting and annotating such real handwriting data is expensive, time-consuming, and often limited by privacy or copyright constraints. This data scarcity motivates the use of large-scale synthetic datasets, which can be generated cheaply and in virtually unlimited quantities. Yet, while synthetic data capture broad visual diversity, they fail to reproduce the subtle statistics of real handwriting, such as its texture, irregularities, and stylistic noise. As a result, models trained solely on synthetic data exhibit a pronounced \textit{synthetic-to-real} gap~\cite{garrido_cvpr_2025}. Closing this gap requires fine-tuning with real samples from the target domain, an assumption that breaks down in many practical settings, where no or just a few annotated real data are available.

To overcome this limitation, we take a different perspective: instead of adapting models with real data, we transfer the synthetic-to-real correction learned when real data is available. Specifically, we address \textbf{zero-shot synthetic-to-real generalization} in HTR:
\vspace{-0.1em}
\begin{quote}
\textit{How can a model trained only on synthetic data perform effectively on real handwriting, without ever seeing a target-domain example?}
\end{quote}
\vspace{-0.1em}
Our approach builds upon the principles of \emph{task arithmetic}~\cite{ilharco2023task}, which represents model fine-tuning as a vector displacement in parameter space, and explores how these task vectors can be combined to induce new behaviors. In particular, we leverage \emph{task analogies}, where relationships between tasks can be expressed as \quotationmarks{$A$ is to $B$ as $C$ is to $D$}. In our setting, this analogy translates naturally to handwriting domains: a model fine-tuned on synthetic data from a source language ($S$) can be related to its real-data counterpart, and this relationship can be transferred to a target language ($T$). 
Moreover, we propose to consider \textbf{multiple analogies}, over different source languages, and then to combine them weighted based on a language-aware similarity score between the source language and the target language. 
By estimating the synthetic-to-real shift from such source domains and applying it to the synthetic model of the target language, we can approximate its real-domain equivalent, thus enabling adaptation to real handwriting entirely through source-derived analogies, without any target-domain samples. 

We validate our approach across five Latin-script languages and six distinct architectures, spanning classical lightweight CNN-RNN models \cite{multidimensional_recurrent_puig_2017, Coquenet:TPAMI:2023} and large-scale Transformer models~\cite{trocr_li_2023,endtoend_coquenet_2022,li_htr-vt_2025}. 
Across all configurations, our method mitigates the synthetic-to-real gap and exhibits strong cross-lingual generalization.

Our main contributions are threefold:
\begin{itemize}
    \item We introduce a novel \textbf{zero-shot task analogy framework} for HTR that transfers the synthetic-to-real adaptation learned from source domains to target domains without requiring any real handwriting data.
    \item We extend this framework to a \textbf{multilingual formulation}, enabling the combination of multiple source analogies through language similarity-based weighting.
    \item We provide a systematic analysis of how synthetic-to-real corrections transfer across languages, showing both a \textbf{strong language-specific component} and a robust \textbf{language-agnostic effect} that improves unrelated languages. Additionally, linear probing reveals that analogy merging produces \textbf{better-aligned representations}.
\end{itemize}
\section{Related Work}
\label{sec:related-work}

\subsection{Handwritten Text Recognition}
\label{sec:related-word-htr}
Bidirectional Long Short-Term Memory networks~\cite{Bi-LSTMGRAVES2005602, Hochreiter-LSTM} trained with the Connectionist Temporal Classification (CTC) objective~\cite{connectionist_graves_2006} dominated HTR benchmarks~\cite{boosting_aradillas_2021, icdar2017_snchez_2017, icfhr2014_snchez_2014, international_abed_2010}. In recent years, attention-based encoder-decoder models~\cite{Bahdanau:ICLR:2015} have emerged as strong alternatives, offering competitive performance across several studies~\cite{attentionhtr_kass_2022, lexicon_kumari_2022, attentionbased_abdallah_2020, endtoend_coquenet_2022}. A comparative analysis in ~\cite{evaluating_michael_2019} provides a detailed account of these sequence-to-sequence approaches.
Transformer architectures~\cite{TransformerVaswani} and their vision extensions~\cite{Dosovitskiy2020AnII} further reshaped the field, enabling more scalable and parallelizable HTR systems~\cite{training_barrere_2024, weakly_paul_2023, improving_sang_2019, dtrocr_fujitake_2023, rethinking_diaz_2021, characterbased_poulos_2021}. These models are typically employed either in encoder-decoder configurations~\cite{trocr_li_2023, transformerbased_momeni_2023, ocformer_mostafa_2021, transformer_wick_2021} or combined with a CTC objective~\cite{rescoring_wick_2021, dan_coquenet_2023, light_barrere_2022}. Diaz \etal~\cite{rethinking_diaz_2021} identified hybrid architectures pairing a convolutional backbone with a Transformer encoder and a CTC decoder as particularly effective.
Transformer-based systems, however, depend heavily on large labeled corpora for pre-training~\cite{trocr_li_2023, rethinking_diaz_2021, vilbert_lu_2019, xlnet_yang_2019, simpler_lu_2021}. Recent studies suggest that self-supervised learning can learn robust representations leveraging unlabeled data~\cite{reading_yang_2022, selfsupervised_pearrubia_2024, sequencetosequence_aberdam_2021}.

\subsection{Adaptation and Generalization in HTR}
Transfer Learning (TL) and Domain Adaptation (DA) are established strategies for improving HTR performance on small datasets \cite{survey_weiss_2016, comprehensive_zhuang_2019, survey_tan_2018}. Early research showed that retraining specific layers or combining TL with DA yields the best results, though TL alone achieves comparable performance \cite{aradillas_jaramillo_boosting_2018, aradillas_boosting_2021}. Fine-tuning remains a strong DA baseline, particularly with CTC-based architectures \cite{kohut_towards_2023}. Other work has focused on pre-training using synthetic data or data generated via Handwritten Text Generation (HTG) \cite{handwritten_pippi_2023, vatr_vanherle_2024}, often evaluating scenarios where target language or author information is known \cite{pippi_how_2023}.


The field closest to our work is Unsupervised Domain Adaptation (UDA) \cite{deep_wang_2018, survey_wilson_2018}, known as Writer Adaptation (WA) in HTR, which adapts models to unseen writers \cite{zhang_sequence--sequence_2019, soullard_improving_2019, domain_zhou_2021, van_der_werff_writer_2023, tula_is_2023} or adapts from synthetic to real data \cite{kang_unsupervised_2020}. Critically, UDA relies on accessing unlabeled target domain samples. In contrast, our research aligns with Domain Generalization (DG) \cite{domain_zhou_2021, in_search_lost_dg_2021, generalizing_wang_2021, learning_wang_2021, learning_qiao_2020}, where no data of any type is available from the OOD target domain. Recent studies in HTR \cite{garrido_cvpr_2025} concluded that existing architectures provided poor DG results, highlighting that linguistic divergence is the major factor affecting generalization. However, this work did not propose a direct method to address this performance gap, which is the specific contribution and focus of our current research.


\subsection{Model Merging}
\label{sec:related-work-task-analogies}
Model merging refers to the process of combining the knowledge of multiple fine-tuned models into a single set of parameters without joint training~\cite{matena2021merging,wortsman2022robust}. A key breakthrough came with Task Arithmetic~\cite{ilharco2023task}, which introduced the idea that the parameter shift induced by fine-tuning can be added, subtracted, or composed linearly to induce new capabilities~\cite{ilharco2023task,wortsman2022robust,panariello2025accurate,yang2024model,yadav2023ties,marouf2023weighted,marczak2024magmax,marczak2025task,tsv}. This perspective naturally extends to task analogies, where relations of the form \quotationmarks{$A$ is to $B$ as $C$ is to $D$} become linear expressions in parameter space. Recent work has demonstrated that this analogy-based formulation can effectively bridge domain gaps, including the synthetic-to-real mismatch in automatic speech recognition~\cite{su2024task}, providing compelling evidence that parameter-space analogies capture transferable structure across domains.
%
%
Our method aligns with this line of work by leveraging task-vector analogies rather than full model merging, exploiting source-domain synthetic-to-real task vectors as analogy components, enabling us to predict target-domain real-data shifts without requiring any real handwriting samples.

\section{Methodology}\label{sec:methodology}
Generalizing HTR models trained on synthetic datasets to real-world data is challenging due to the inherent visual and linguistic domain shift~\cite{bhunia_metahtr_2021,garrido2025htrsurvey,garrido_cvpr_2025}. To address this issue, we aim to transfer known synthetic-to-real transformations from a source language $S$ to a target language $T$, enabling us to approximate real-domain performance for $T$ without requiring any real handwritten samples.

\subsection{Training the Ancestor Model}
A key requirement for performing task arithmetic is that all task-specific models originate from a shared initialization that encodes broad, domain-general knowledge. In vision tasks this role is often played by large pretrained models such as CLIP~\cite{radford2021learning}; however, no analogous multilingual handwriting model exists. For this reason, we explicitly construct and train our own \emph{ancestor model}, designed to serve as the common starting point from which all subsequent fine-tuned models branch.

To obtain such ancestor model, we begin by building a large multilingual synthetic corpus spanning five Latin-alphabet languages: English (en), French (fr), Italian (it), Spanish (es), and German (de). For each language, we extract text from the WIT-corpus~\cite{wit_dataset}, chosen for its wide coverage and linguistic diversity, and synthesize 25M handwriting samples per language using a pool of \num{3700} publicly available handwritten fonts.\footnote{Although our study focuses on Latin-script languages, the procedure naturally extends to other writing systems by generating additional synthetic data with appropriate fonts.} From this corpus, we generate two dataset variants. The \emph{ancestor} model is first trained on a simplified version of the synthetic data (monochrome images without backgrounds or transformations) to capture general linguistic and handwriting patterns shared across all five languages. This shared multilingual ancestor provides a common reference point that makes subsequent task-vector operations well-defined and effective (see \cref{sec:task_arithmetic}).

Starting from this ancestor, we then train a \emph{child} model for each language on a heavily augmented version of the dataset, including realistic backgrounds and a wide array of geometric and photometric transformations (rotation, blurring, elastic distortions, random perspective). This yields domain-specialized models for each language while strictly preserving the shared initialization required for meaningful task analogies. Further details are provided in \cref{sec:experiments}.

\subsection{Task Vectors and Task Arithmetic}
Given an ancestor model $\theta_0$ and its fine-tuned counterpart $\theta^\star$ on a downstream task, we define the \textit{task vector} as the parameter-space difference $\tau = \theta^\star - \theta_0$, which captures how the base model adapts to the target task. The task arithmetic framework~\cite{ilharco2023task} explores how such vectors can be combined through simple linear operations to induce new behaviors. Adding multiple task vector yields a composite representation that integrates the capabilities of all included tasks, while subtracting a task vector can effectively remove the knowledge associated with that task. Task analogies extend this idea by leveraging the relationships of the form \quotationmarks{$A$ is to $B$ as $C$ is to $D$}, to enhance performance on new tasks with limited or no labeled data. The corresponding task vector composition follows:
\begin{equation}
    \tau_{\text{new}} = \tau_C + (\tau_B - \tau_A),
\end{equation}
where $\tau_{\text{new}}$ is the vector that transfers to task $C$ the transformation that maps $A$ to $B$, enabling the model to approximate task $D$ without accessing its data. In handwriting recognition, synthetic-to-real shifts tend to be consistent across languages (\eg, stroke noise, texture, variability), making this analogy formulation particularly effective for transferring real-data corrections. In this work, we leverage this analogy mechanism to obtain competitive models on previously unseen real-world handwriting data, directly addressing out-of-distribution generalization in HTR only using synthetic target pretrained models.

\begin{figure}[t]
    \centering
    \includegraphics[width=\linewidth]{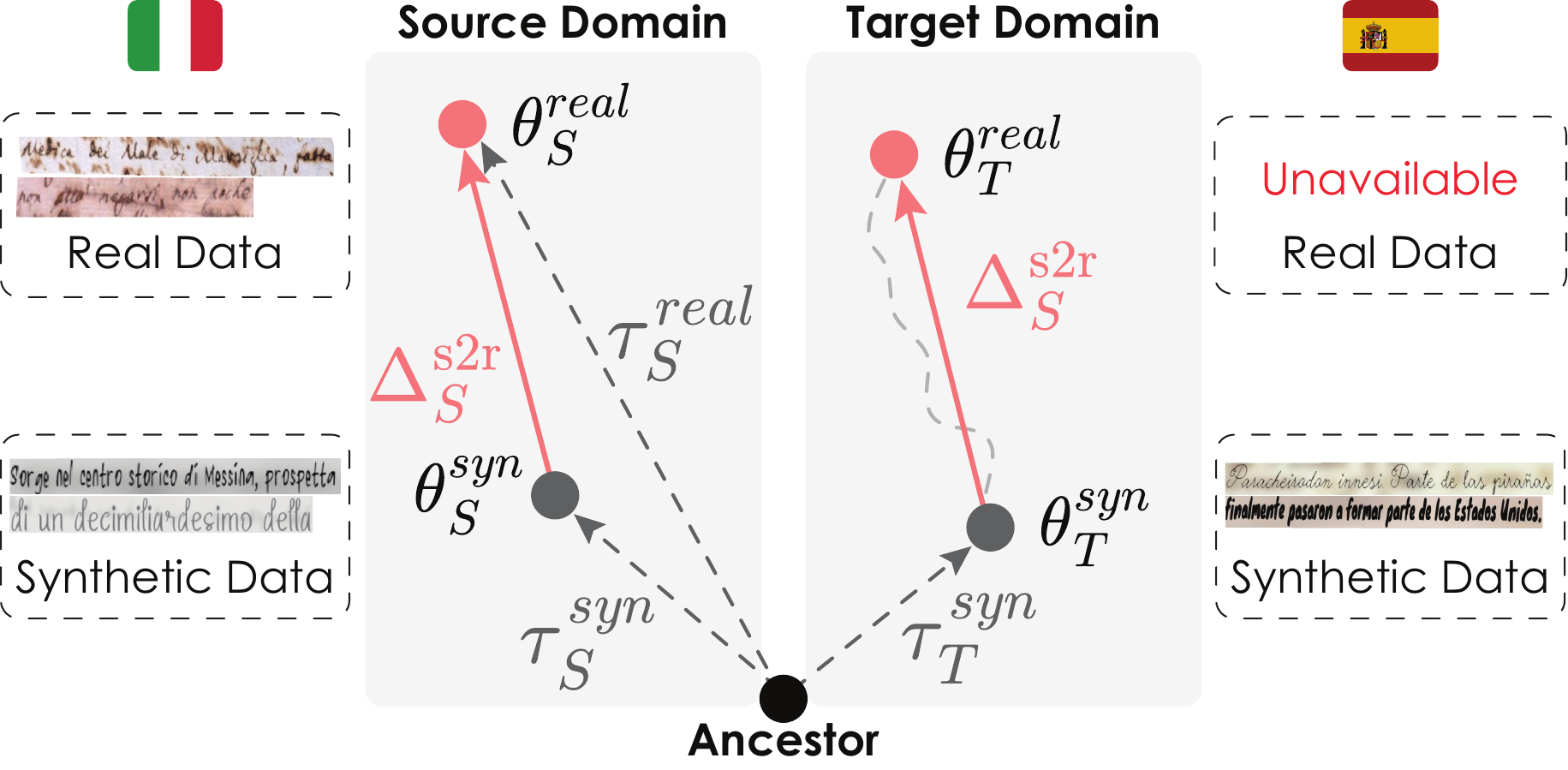}
    \caption{Example of Single Analogy Merging: Estimating the Spanish real-domain model ($\theta_{\text{es}}^{\text{real}}$) by transferring the synthetic-to-real displacement vector ($\Delta_{\text{it}}^{\text{s2r}}$) learned from the Italian domain.}
    \label{fig:analogies}
\end{figure}
\subsection{Task analogies}
\label{sec:task_arithmetic}
We propose two variants of task analogies: (i) a \textit{monolingual analogy}, which relies on a single source language, and (ii) a \textit{multilingual analogy}, which merges information from multiple sources using heuristic importance weights.

\tit{Monolingual analogy.} Let $\theta_0$ be the ancestor model, and let $\theta_S^{syn}$ and $\theta_T^{syn}$ be the models obtained by fine-tuning $\theta_0$ on synthetic data from the source and target language, respectively. Further fine-tuning $\theta_S^{syn}$ on real data from $S$ yields $\theta_S^{real}$. For the source domain, we define the corresponding task vectors:
\begin{equation}
\tau_S^{\text{syn}} = \theta_S^{\text{syn}} - \theta_0, \quad
\tau_S^{\text{real}} = \theta_S^{\text{real}} - \theta_0.
\end{equation}
Here, $\tau_S^{\text{syn}}$ encodes the adaptation from the base model to the synthetic source data, while $\tau_S^{\text{real}}$ captures the corresponding adaptation to real data. Their difference,
\begin{equation}
\Delta_S^{\text{s2r}} = \tau_S^{\text{real}} - \tau_S^{\text{syn}},
\end{equation}
represents the synthetic-to-real transformation learned from the source domain. Following the task analogy principle, \ie, $\tau_S^{\text{syn}}$ is to $\tau_S^{\text{real}}$ as $\tau_T^{\text{syn}}$ is to $\tau_T^{\text{real}}$, we transfer the synthetic-to-real displacement to the target domain:
\begin{equation}
\label{eq:analogy}
\theta_T^{\text{real}} \approx \theta_0 + \tau_T^{\text{syn}} + (\tau_S^{\text{real}} - \tau_S^{\text{syn}}) = \theta_T^{\text{syn}} + \Delta_S^{\text{s2r}},
\end{equation}
\begin{figure*}[t]
    \centering
    \includegraphics[width=\linewidth]{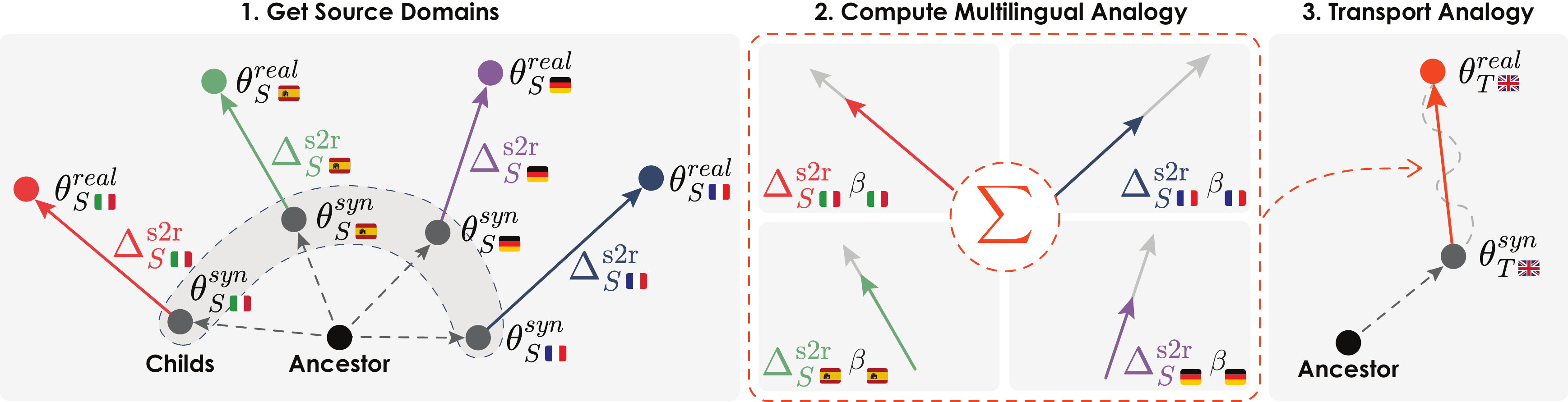}
    \caption{Multilingual analogy merging framework for Zero-Shot Synthetic-to-Real HTR. We estimate the target real model $\theta_T^{\text{real}}$ by combining synthetic-to-real displacement vectors $\Delta_S^{\text{s2r}}$ from multiple sources $\mathcal{S}$. These vectors are weighted by linguistically informed coefficients $\beta_s$ derived from source language $S$ and synthetic target $T$ similarity, achieving zero-shot adaptation (\cref{eq:multi-analogy}).}
    \label{fig:fig_methodology}
\end{figure*}

which provides an estimate of the target’s real-domain parameters without requiring any real data. A graphical representation is shown in \cref{fig:analogies}.

To improve stability and control the magnitude of the transferred displacement, following~\cite{wortsman2022robust}, we apply a linear interpolation between the base model $\theta_T^{\text{syn}}$ and the analogy-based estimate in~\cref{eq:analogy}, yielding:
\begin{equation}\label{eq:wise-ft}
\theta_T^{\text{real}} \approx (1-\alpha)\, \theta_T^{\text{syn}} + \alpha \, (\theta_T^{\text{syn}} + \Delta_S^{\text{s2r}}) = \theta_T^{\text{syn}} + \alpha \, \Delta_S^{\text{s2r}}.
\end{equation}
The scaling coefficient $\alpha$ controls the strength of the transferred displacement from source to target while preserving the structure of the analogy. Specifically, we choose $\alpha$ using only real data from non-target languages (heldout selection \cite{in_search_lost_dg_2021}), ensuring that model selection remains fully zero-shot with respect to the target domain. A detailed description of this procedure is provided in \cref{sec:experiments}.

\tit{Multilingual analogy.} The single source analogy provides the transformation from only one source language, but it does not exploit the rich cross-lingual structure available when multiple source domains are present.  In practice, each source $s$ provides a different estimate of the synthetic-to-real displacement $\Delta_s^{\text{s2r}}$, capturing language-specific visual and statistical shifts. Relying on a single source may therefore introduce bias or limit generalization.

To exploit the available knowledge, we extend task analogies to a multilingual formulation. Let $\mathcal{S}$ be the set of available source languages. For each source $s \in \mathcal{S}$, we compute its synth-to-real task vector $\Delta_s^{\text{s2r}}$. Building on the additive nature of task arithmetic, we combine these source vectors using a weighted linear superposition. Each weight term determines how strongly a given source contributes to the final target estimate.

Formally, for a target language $T$, we define an \textbf{importance coefficient} $\beta_{(\cdot)}(s,T)$ for each source, that quantifies how informative the source $s$ is for the target $T$. This coefficient, may come from any similarity score (noted by ${(\cdot)}$), we will discuss different strategies in \cref{sec:importance_coeff}. Intuitively, these coefficients modulate the contribution of each source analogy: sources linguistically closer to the target (\eg, sharing character sets or $n$-gram statistics) should have larger weights. This leads to the multilingual task analogy:
\begin{equation}\label{eq:multi-analogy}
\theta_T^{\text{real}}
\approx
\theta_T^{\text{syn}}
+
\alpha \sum\nolimits_{s \in \mathcal{S}} \beta_{(\cdot)}(s,T) \, \Delta_s^{\text{s2r}}.
\end{equation}
We refer to this process as multilingual analogy, since the target model is constructed by merging the synth-to-real displacements of multiple sources. In the following subsection, we detail how the coefficients $\beta_{(\cdot)}(s,T)$ are computed from the synthetic corpora using various linguistic similarity metrics. These scores allow multilingual analogy to selectively combine source transformations in a principled manner without employing handwritten data. The multilingual analogy process is depicted in Fig. \ref{fig:fig_methodology}.

\subsection{Importance Coefficient}\label{sec:importance_coeff}
A critical challenge in the multilingual analogy formulation of \cref{eq:multi-analogy} lies in the selection of the weights $\beta_s$. As the number of potential sources increases, the weight space becomes high-dimensional, and direct optimization of $\beta_s$ is infeasible. Therefore, we compute $\beta_s$ using similarity measures computed solely from the synthetic corpora, ensuring applicability in zero-data scenarios without requiring any real handwritten sample from the target domain.

Specifically, we compute the frequency distribution of character $n$-grams for each language using the synthetic validation corpus with $n \in N={1,\dots, 5}$. We then assess pairwise similarities between languages using three strategies.
\tit{1. KL Divergence.} We compute the (average) KL divergence between $S$ and $T$ across all $n$-gram orders:
\begin{equation}
D_{\mathrm{KL}}(S,T)
= \frac{1}{N}\sum_{n=1}^{N} D_{\mathrm{KL}}^{(n)}(S\,\|\,T),
\end{equation}
since KL divergence is unbounded, we obtain its similarity score version, bounded between $[0,1]$, via:
\begin{equation}
    \beta_{\mathrm{KL}}(S,T)
= 1 - \frac{D_{\mathrm{KL}}(S,T)}{\max_{(i,j)} D_{\mathrm{KL}}(i,j)}.
\end{equation}
\tinytit{2. Hellinger Distance.} We concatenate the $n$-gram distributions in a single vector $\mathbf{V} = [P_{1}, P_{2}, \dots, P_{5}]$, and, similarly as in (1), we obtain the bounded similarity version as:
\begin{equation}
\beta_{H}(S,T)
= 1 - \frac{1}{\sqrt{2}} \,\bigl\|\sqrt{\mathbf{V}_S}-\sqrt{\mathbf{V}_T}\bigr\|_2.
\end{equation}
\tinytit{3. Jaccard Similarity (IoU).} Let $\mathcal{V}_S$ and $\mathcal{V}_T$ be the sets of unique $n$-grams from each language, IoU is computed as:
\begin{equation}
    \beta_{J}(S,T)
= \frac{|\mathcal{V}_S \cap \mathcal{V}_T|}
       {|\mathcal{V}_S \cup \mathcal{V}_T|}.
\end{equation}
These similarity scores provide complementary views of the pairwise relationship of the different languages, and lead to different weighting behaviors in \cref{eq:multi-analogy}. KL Divergence focuses on distributional alignment, emphasizing whether two languages exhibit similar $n$-gram frequency patterns. Hellinger Distance, being symmetric and geometry-aware, captures global statistical proximity between languages while being less sensitive to rare $n$-grams than KL. In contrast, Jaccard similarity ignores frequencies altogether and measures only the lexical overlap between languages, \ie, whether they share the same character sequences, regardless of how often they occur.


This provides a richer, more robust mechanism for assigning importance coefficients $\beta_s$ in multilingual task analogies, reflecting different facets of linguistic similarity.

\section{Experiments}\label{sec:experiments}
\begin{table*}[t]
    \caption{Average Character Error Rate (CER $\downarrow$) and Word Error Rate (WER $\downarrow$) across all real handwritten datasets for six HTR architectures. Performance is compared against the \textbf{In-domain upper bound} and the \textbf{Baseline} ($\theta_T^{\text{syn}}$) zero-shot model. Results are grouped by single analogy and multiple analogy strategies, incorporating uniform and linguistically informed ($\beta_{KL}, \beta_{H}, \beta_{J}$) priors for zero-shot adaptation. Values in parentheses denote the absolute improvement ($\Delta$) achieved relative to the Baseline.}
    \centering
    \setlength{\tabcolsep}{.3em}
    \resizebox{\linewidth}{!}{%
    \footnotesize
    \begin{tabular}{l l ll l ll l ll l ll l ll l ll}
    \toprule
    && \multicolumn{2}{c}{\makecell{\textbf{TrOCR (S)} \cite{trocr_li_2023}}} 
    && \multicolumn{2}{c}{\makecell{\textbf{TrOCR (B)} \cite{trocr_li_2023}}} 
    && \multicolumn{2}{c}{\makecell{\textbf{TrOCR (L)} \cite{trocr_li_2023}}}
    && \multicolumn{2}{c}{\makecell{\textbf{CRNN} \cite{multidimensional_recurrent_puig_2017}}}
    && \multicolumn{2}{c}{\makecell{\textbf{VAN} \cite{endtoend_coquenet_2022}}}
    && \multicolumn{2}{c}{\makecell{\textbf{HTR-VT} \cite{li_htr-vt_2025}}} \\
    \cmidrule{3-4} \cmidrule{6-7} \cmidrule{9-10} \cmidrule{12-13} \cmidrule{15-16} \cmidrule{18-19}
    && \textbf{CER} $\downarrow$ & \textbf{WER} $\downarrow$ && \textbf{CER} $\downarrow$ & \textbf{WER} $\downarrow$ && \textbf{CER} $\downarrow$ & \textbf{WER} $\downarrow$ && \textbf{CER} $\downarrow$ & \textbf{WER} $\downarrow$ && \textbf{CER} $\downarrow$ & \textbf{WER} $\downarrow$ && \textbf{CER} $\downarrow$ & \textbf{WER} $\downarrow$ \\
    \midrule
    In-domain && 5.0& 16.7&& 4.4& 14.8&& 4.3& 14.1&& 4.2& 16.0&& 5.4& 21.8&& 5.3& 21.8\\
    Baseline $\theta_T^{syn}$ && 39.7\deltazero & 72.4\deltazero&& 37.7\deltazero& 69.1\deltazero&& 36.2\deltazero& 67.8\deltazero&& 38.8\deltazero& 75.6\deltazero&& 35.2\deltazero& 76.3\deltazero&& 41.7\deltazero& 81.8\deltazero\\
    \midrule
    \rowcolor{gray!20}
    \multicolumn{3}{l}{\textbf{Single analogy}} &&& & && & && & && & && & \\
    \midrule
    $\beta=1$ &&     $37.9\deltapos{-1.8}$& $72.8\deltaneg{+0.4}$&& $36.9\deltapos{-0.8}$& $70.9\deltaneg{+1.8}$&& $35.3\deltapos{-0.9}$& $73.1\deltaneg{+5.3}$&& $35.9\deltapos{-2.9}$& $73.3\deltapos{-2.2}$&& $34.5\deltapos{-0.7}$& $78.0\deltaneg{+1.7}$&& $40.6\deltapos{-1.1}$& $83.7\deltaneg{+1.9}$\\
    \midrule
    $\beta = \beta_{\text{KL}}$  && $35.0\deltapos{-4.7}$& $68.2\deltapos{-4.2}$&& $32.9\deltapos{-4.8}$& $64.6\deltapos{-4.5}$&& $32.4\deltapos{-3.8}$& $64.5\deltapos{-3.3}$&& $35.4\deltapos{-3.4}$& $70.0\deltapos{-5.6}$&& $31.7\deltapos{-3.5}$& $72.1\deltapos{-4.2}$&&$36.6\deltapos{-5.1}$&$77.1\deltapos{-4.7}$\\
    $\beta = \beta_{\text{H}}$  && $35.3\deltapos{-4.4}$& $68.7\deltapos{-3.7}$&& $33.0\deltapos{-4.7}$& $64.9\deltapos{-4.2}$&& $30.9\deltapos{-5.3}$& $63.6\deltapos{-4.2}$&& $35.3\deltapos{-3.5}$& $71.2\deltapos{-4.4}$&& $31.7\deltapos{-3.5}$& $72.3\deltapos{-4.0}$&&$36.8\deltapos{-4.9}$&$77.4\deltapos{-4.4}$\\
    $\beta = \beta_{\text{J}}$ && $35.5\deltapos{-4.2}$& $69.2\deltapos{-3.2}$&& $33.7\deltapos{-4.0}$& $65.3\deltapos{-3.8}$&& $31.3\deltapos{-4.9}$& $64.6\deltapos{-3.2}$&& $35.9\deltapos{-2.9}$& $72.4\deltapos{-3.2}$&& $32.6\deltapos{-2.6}$& $73.8\deltapos{-2.5}$&&$37.7\deltapos{-4.0}$&$79.0\deltapos{-2.8}$\\
    \midrule
    \rowcolor{gray!20}
    \multicolumn{3}{l}{\textbf{Multiple analogy}} & && & && & && & && & && & \\
    \midrule
    $\beta=1$ && $54.8\deltaneg{+15.1}$& $78.9\deltaneg{+6.5}$&& $33.8\deltapos{-3.9}$& $67.1\deltapos{-2.0}$&& $39.1\deltaneg{+2.9}$& $73.9\deltaneg{+6.1}$&& $51.4\deltaneg{+12.6}$& $76.0\deltaneg{+0.4}$&& $30.2\deltapos{-5.0}$& $70.3\deltapos{-6.0}$&&$33.7\deltapos{-8.0}$&$74.2\deltapos{-7.6}$\\
    $\beta=1/N$ && $35.3\deltapos{-4.4}$& $73.1\deltaneg{+0.7}$&& $31.8\deltapos{-5.9}$& $64.2\deltapos{-4.9}$&& $31.5\deltapos{-4.7}$& $64.9\deltapos{-2.9}$&& $34.8\deltapos{-4.0}$& $70.6\deltapos{-5.0}$&& $30.2\deltapos{-5.0}$& $70.3\deltapos{-6.0}$&&$34.1\deltapos{-7.6}$&$75.4\deltapos{-6.4}$\\
    \midrule
    $\beta = \beta_{\text{KL}}$  && $35.0\deltapos{-4.7}$& $68.3\deltapos{-4.1}$&& $32.6\deltapos{-5.1}$& $63.9\deltapos{-5.2}$&& $32.4\deltapos{-3.8}$& $64.7\deltapos{-3.1}$&& $35.0\deltapos{-3.8}$& $70.8\deltapos{-4.8}$&& $31.0\deltapos{-4.2}$& $71.1\deltapos{-5.2}$&&$34.1\deltapos{-7.6}$&$74.0\deltapos{-7.8}$\\
    $\beta = \beta_{\text{H}}$  && $34.4\deltapos{-5.3}$& $68.3\deltapos{-4.1}$&& $32.5\deltapos{-5.2}$& $65.2\deltapos{-3.9}$&& $30.9\deltapos{-5.3}$& $63.6\deltapos{-4.2}$&& $35.1\deltapos{-3.7}$& $70.7\deltapos{-4.9}$&& $31.0\deltapos{-4.2}$& $71.1\deltapos{-5.2}$&&$33.6\deltapos{-8.1}$& $73.9\deltapos{-7.9}$\\
    $\beta = \beta_{\text{J}}$ && $33.7\deltapos{-6.0}$& $67.1\deltapos{-5.3}$&& $32.1\deltapos{-5.6}$& $64.4\deltapos{-4.7}$&& $31.3\deltapos{-4.9}$& $64.6\deltapos{-3.2}$&& $34.7\deltapos{-4.1}$& $70.5\deltapos{-5.1}$&& $29.9\deltapos{-5.3}$& $69.8\deltapos{-6.5}$&&$34.0\deltapos{-7.7}$&$74.7\deltapos{-7.1}$\\
    \bottomrule
    \end{tabular}}

\label{tab:main_results}
\end{table*}

\subsection{Experimental setup}
\tit{Real data.}We evaluated the performance of our approach in the
following real handwritten datasets: IAM \cite{iamdatabase_marti_2002} (English), Rimes \cite{rimes_2010} (French), LAM \cite{cascianelli2022lam} (Italian), Rodrigo \cite{serrano-etal-2010-rodrigo} (Spanish) and ICFHR 2016 (READ 2016) \cite{icfhr_2016_sanchez} (German). 
Original data splits for each of these datasets were maintained. Further details are reported in the supplementary material.
%

\tit{HTR models.}
Our experimental setup encompasses six distinct HTR model architectures, ranging in size from 1.7M to 500M parameters. Specifically, we include classical HTR literature models based on Convolutional and Recurrent Networks, such as CRNN \cite{multidimensional_recurrent_puig_2017} and VAN \cite{Coquenet:TPAMI:2023}, both trained with the CTC \cite{connectionist_graves_2006} objective. We also incorporate recent state-of-the art Transformer-based models, namely the TrOCR \cite{trocr_li_2023} architecture across its three canonical sizes: $\text{TrOCR}\ (\text{S})$, $\text{TrOCR}\ (\text{B})$, and $\text{TrOCR}\ (\text{L})$, which utilize an autoregressive decoding mechanism. Additionally, we include a recent hybrid ViT model trained with the CTC objective, HTR-VT \cite{li_htr-vt_2025}. More details about the architectures are provided in the supplementary material.

\tit{Implementation details.}
We strictly followed the original training configurations of each architecture, including hyperparameters such as batch size, optimizers, and warm-up schedules. For a fair comparison, all models are trained from scratch. For the synthetic data training stage, models are trained for 16M steps, saving the best model based on synthetic validation performance every 160K steps. Models trained on real HTR datasets were trained for 100K steps, saving the model that achieves the best performance on that specific real dataset's validation split. We applied the same set of data augmentation and transformations (detailed in the supplementary material), are applied to both the synthetic datasets (used for training the child models) and the real HTR datasets.

\tit{Selection of the scaling coefficient.} In our experimental protocol, the scaling coefficient $\alpha$ is selected without using any real data from the target domain to maintain the zero-shot setting. Specifically, we determine $\alpha$ using a held-out of real datasets drawn only from non-target languages. For each merging configuration, we run a grid search over $\alpha \in [0, 1]$ with a step size of $0.125$ and compute the average CER across the held-out languages. The selected $\alpha$ is the one yielding the lowest average CER. Once selected, the same $\alpha$ is used for all target evaluations under the same configuration, ensuring that model selection remains independent of the target language. A detailed analysis of the robustness of this selection strategy is provided in \cref{sec:alpha_robustness}.

\subsection{Results}\label{sec:main_results}
\Cref{tab:main_results} summarizes the zero-shot synthetic-to-real performance of our approach. Specifically, we report the average Character Error Rate (CER) and Word Error Rate (WER) across all real handwritten datasets for the six architectures considered. We establish two reference points to contextualize the results: the \textit{In-domain} setting, representing the upper bound obtained through fully supervised training on real data, and the \textit{Baseline} ($\theta_T^{\text{syn}}$), \ie, the model trained solely on synthetic data and evaluated on the real-data datasets. This setting reflects the synthetic-to-real gap under a zero-shot scenario with no analogy strategy. We evaluate the two analogy-based strategies introduced in \cref{sec:methodology}. The \textit{single-analogy} case applies the synthetic-to-real displacement from one source language, while the \textit{multi-analogy} case merges the displacements from multiple sources. For clarity, we treat the single-analogy setup as a special instance of the multi-analogy formulation in \cref{eq:multi-analogy}, where the source set reduces to a singleton (\(|\mathcal{S}|=1\)). Both strategies are evaluated in two variants a \textit{non-language-informed} variant, and a \textit{language-informed} variant.

A \textbf{non-language-informed} variant is one in which all sources are treated equally, without using any linguistic information about the target. In the single-analogy case, this corresponds to selecting the source that performs best on the held-out real datasets (with $\beta = 1$ for that source). In the multi-analogy case, the analogue is to assign equal importance to all sources, either uniformly ($\beta = 1$) or normalized ($\beta = 1/N$). Conversely, a \textbf{language-informed} variant incorporates linguistic proximity between each source $s$ and the target $T$ through the similarity-based importance coefficients introduced in \cref{sec:importance_coeff}. These coefficients, modulate the contribution of each $\Delta_s^{\mathrm{s2r}}$ in proportion to its estimated relevance for the target domain.

Across all architectures, analogy-based merging substantially reduces the synthetic-to-real gap. Single-analogy corrections already yield strong improvements, while multi-analogy merging, especially when guided by linguistic similarity, provides the most consistent zero-shot gains.

\tit{Effect of Single and Multiple Sources.} First, applying the single-analogy synthetic-to-real correction, already yields substantial improvements over the $\theta_T^{\text{syn}}$ Baseline across all evaluated datasets and architectures. This confirms that even a single learned displacement carries meaningful information about the synthetic-to-real shift and can significantly reduce the zero-shot gap. Second, using multiple sources further improves performance. Across models, the multi-analogy formulation achieves larger average reductions in both CER and WER, indicating that aggregating complementary corrections from several languages provides a more robust approximation of the target real-domain model.

\begin{figure}[t]
    \centering
    \includegraphics[width=.95\linewidth]{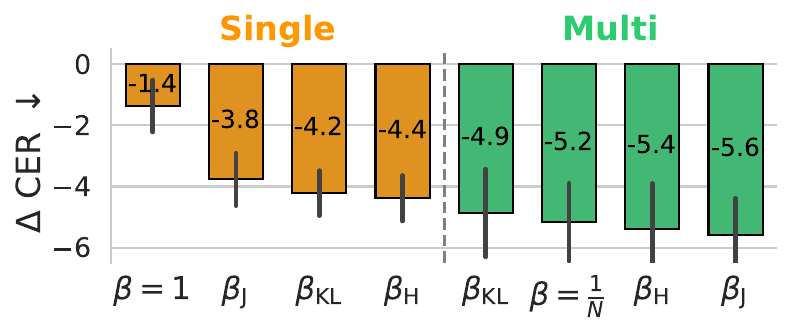}
    \caption{Zero-shot $\Delta\text{CER}$ improvement across all datasets and architectures. We compare single and multi-analogy methods (incorporating linguistic priors) against non-informed baselines (Single $\beta=1$ and $\beta=1/N$ uniform weighting). Linguistically aware multi-analogy methods achieved the best overall gains.}
    \label{fig:aggregated_improvement}
\end{figure}
\begin{figure*}[t]
    \centering
    \includegraphics[width=\linewidth]{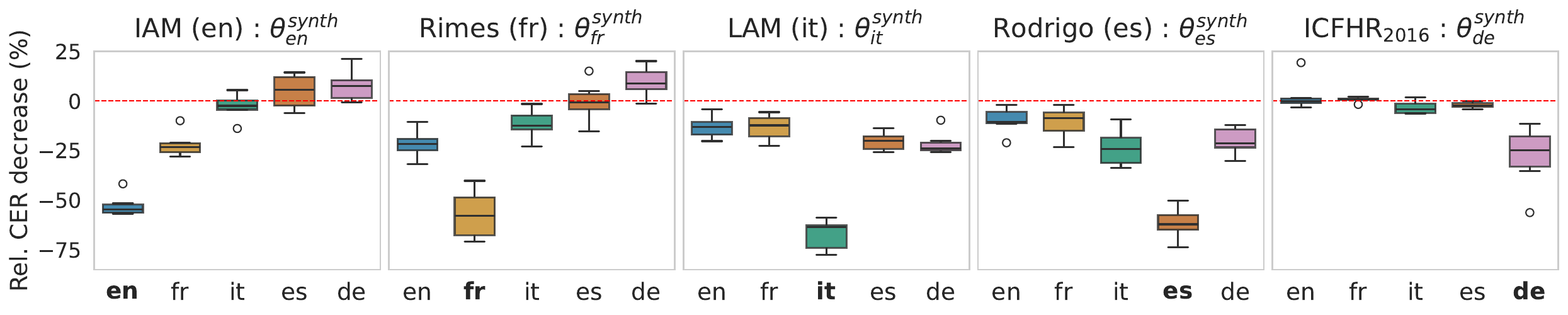}
    \caption{Distribution of Relative CER decrease (\%) achieved using single, unweighted analogies ($\beta=1$) across five target languages ($T$). The largest reduction in error (best transfer) is consistently achieved when the source analogy ($S$) matches the target language ($T$), confirming the strong influence of direct linguistic alignment on the vector's transferability.}
\label{fig:single-analogies-results}
\end{figure*}
\tit{Linguistic Awareness Improves Analogies.} Focusing on the multi-analogy strategies, we observe that the linguistically informed weighting consistently outperforms uniform schemes. In particular, the similarity coefficients $\mathbf{\{\beta_{\mathrm{KL}},\, \beta_{\mathrm{H}},\, \beta_{\mathrm{J}}\}}$, provide more effective combinations of source displacements than both $\beta=1$ and the normalized uniform variant $\beta=1/N$. This supports our central intuition that the usefulness of a source analogy is not uniform: sources that are linguistically closer to the target contribute more reliable synth-to-real corrections.

\Cref{fig:aggregated_improvement} summarizes the average zero-shot CER improvements across all datasets and architectures. The trend is consistent across model families: both the single- and multi-analogy variants that rely on linguistic similarity achieve the largest reductions in error, outperforming the non-informed baselines. Among the proposed metrics, Hellinger distance and Jaccard similarity yield the strongest gains, while KL-based weighting remains competitive but slightly less effective on average.

%
%
%

\paragraph{How far is the optimal scaling factor?}\label{sec:alpha_robustness}
As discussed in~\cite{in_search_lost_dg_2021}, a reliable zero-shot adaptation method must avoid using any real target-domain supervision, that is, its performance should not depend on tuning hyperparameters using target validation data. In \cref{tab:oracle_gap_models}, we compare our fully unsupervised $\alpha$-selection strategy against an \emph{oracle} variant \cite{in_search_lost_dg_2021}, where $\alpha$ is chosen using a real-data validation set from the target domain. This comparison quantifies how close our unsupervised procedure is to the best possible tuning and evaluates the stability of the method under truly zero-shot conditions. The results show that the gap between the held-out and oracle selections remains consistently small across models, indicating that our $\alpha$-selection strategy is robust even without any access to target-domain feedback.

\begin{table}[t]
    \renewcommand{\arraystretch}{0.75}
  \setlength{\tabcolsep}{0.03em}
  \centering
  \captionsetup{font=footnotesize}
  \caption{
  Heldout and oracle performance for all models averaged about methods and datasets.
  Performance obtained by heldout selection remain close  
  to their oracle performance, showing the robustness of our approach.
  }
  \footnotesize
  \renewcommand{\arraystretch}{1.05}
  \begin{tabular}{lccccc}
  \toprule
  Model & CER\textsubscript{oracle} & CER\textsubscript{heldout} & WER\textsubscript{oracle} & WER\textsubscript{heldout} \\
  \midrule
  TrOCR(S)   & $36.2$ & $37.4\deltaneg{+1.2}$  & $67.9$ & $70.0\deltaneg{+2.1}$ \\
  TrOCR(B)   & $31.9$ & $33.3\deltaneg{+1.4}$  & $62.7$ & $65.6\deltaneg{+2.9}$ \\
  TrOCR(L)   & $31.3$ & $32.8\deltaneg{+1.5}$  & $62.3$ & $66.4\deltaneg{+4.1}$ \\
  CRNN       & $36.6$ & $37.0\deltaneg{+0.4}$  & $71.1$ & $71.9\deltaneg{+0.8}$ \\
  VAN        & $30.4$ & $31.4\deltaneg{+1.0}$  & $70.1$ & $72.1\deltaneg{+2.0}$ \\
  HTR-ViT    & $34.6$ & $35.7\deltaneg{+1.1}$  & $74.4$ & $76.6\deltaneg{+2.2}$ \\
  \bottomrule
  \end{tabular}
  \label{tab:oracle_gap_models}
\end{table}

\subsection{Cross-Lingual Transfer Analysis}
To better understand the behavior of the synthetic-to-real displacement vector $\Delta_{S}^{\mathrm{s2r}}$, we analyze the type of information it transfer across languages. In particular, we investigate the extent to which such a displacement reflects \textit{language-specific} adjustments tied to the source $S$, versus a more \textit{language-agnostic} correction that captures general properties of the synthetic-to-real shift. Disentangling these components allows us to assess whether the vector primarily conveys linguistic priors or whether it encodes a broader improvement in synthetic-to-real robustness.

\begin{figure}[t]
    \centering
    \includegraphics[width=\linewidth]{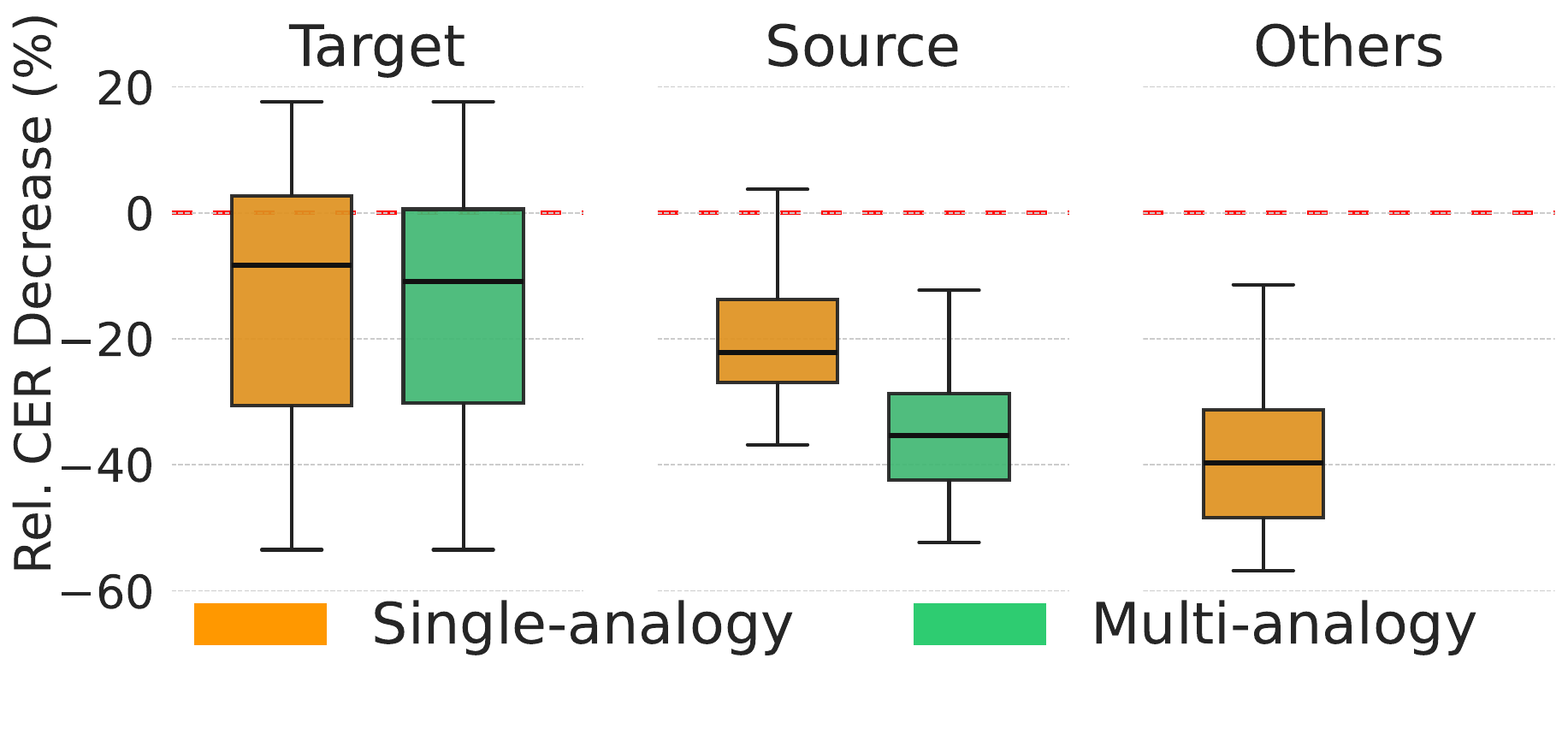}
    \caption{Relative CER improvement across two evaluation scenarios designed for vector bias and generalization. Evaluation on the source domain ($S$) quantifies linguistic bias, while other domain ($Q$) isolates the general language-agnostic component.}
    \label{fig:improvements_type}
\end{figure}

First (\cref{fig:single-analogies-results}), we measure how each source language analogy vector influences all possible target languages, providing a direct view of the cross-lingual sensitivity of analogy transfer. Second (\cref{fig:improvements_type}), we probe the inductive bias within each $\Delta_{S}^{\mathrm{s2r}}$ vectors by comparing its effect on three evaluation settings: (i) the intended target language $T$, (ii) the source language(s) $S$ from which the vector was constructed, and (iii) all remaining languages, defined as the average performance on every dataset that is neither $S$ nor $T$. This third evaluation isolates the general, language-independent component of $\Delta_{S}^{\mathrm{s2r}}$, revealing whether the displacement provides a broadly transferable synthetic-to-real correction beyond the source-target pair.

\begin{figure*}[t]
    \centering
    \includegraphics[width=\linewidth]{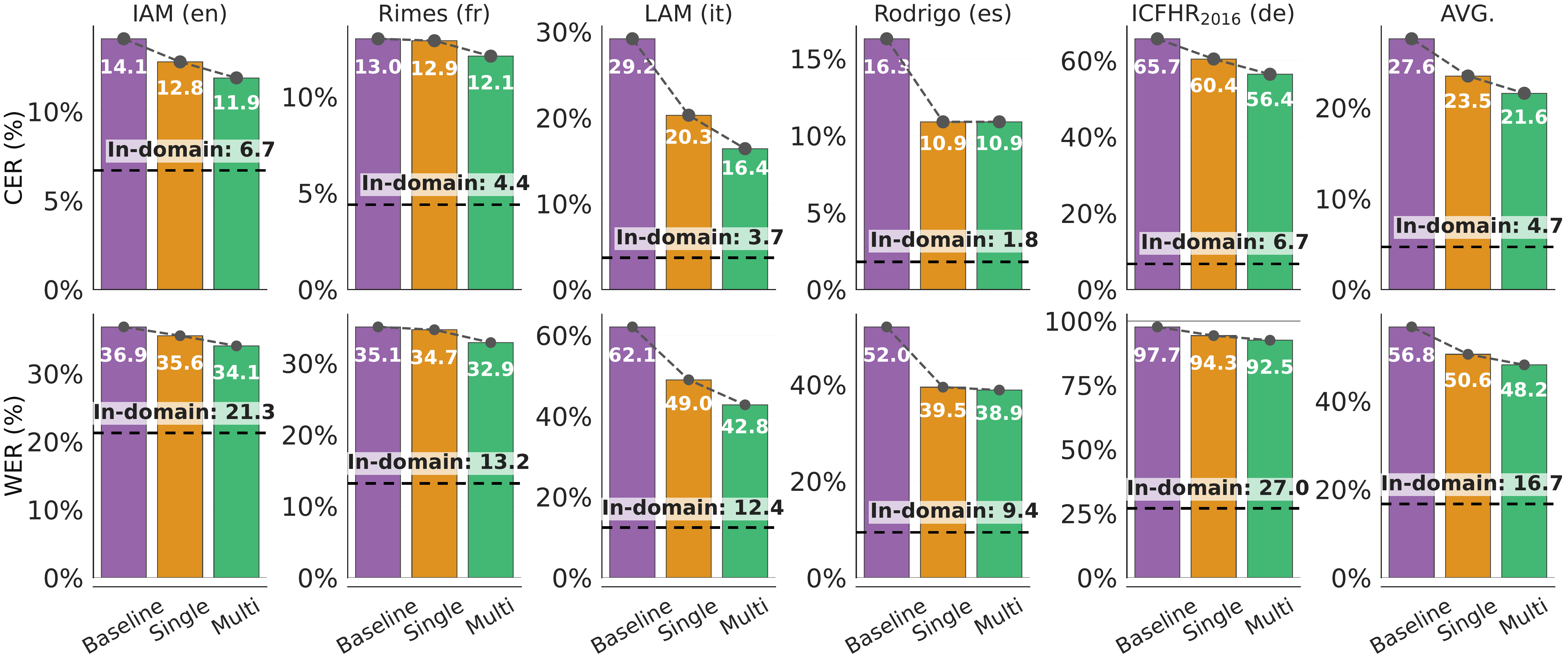}
    \caption{Linear Probing results evaluating representation quality (average CER/WER). We compare the Baseline against the best performing Linguistic Aware Single-Analogy and Multi-Analogy merged models, after fine-tuning only the final linear layer.}
\label{fig:linear_probing}
\end{figure*}

\tit{Language-specific transfer.} We first quantify how well each source language contributes to every possible target through the single-analogy setting. This configuration isolates the transferability of the synthetic-to-real displacement $\Delta_S^{\mathrm{s2r}}$ by measuring its direct effect when applied to a target model $T$. The resulting cross-language transfer results are reported in \cref{fig:single-analogies-results}. Across all six architectures, we observe a consistent and interpretable trend: analogies transfer most effectively when the target language matches the source $S$, indicating that the displacement encodes a substantial language-specific component. Beyond the matching case, improvements decrease as linguistic distance increases. Languages that are closer according to our $n$-gram-based similarity metrics, such as Italian-Spanish or English-German, exhibit stronger reciprocal transfer, whereas more distant pairs show limited benefit. This behavior is stable across architecture families and parameter scales, suggesting that these transfer patterns stem from intrinsic properties of the synth-to-real shift rather than from model-specific artifacts. Refer to the supplementary material for the exact values of the linguistic scores.

\tit{How biased are the synth-to-real vectors?}
To examine what type of information is encoded in each synthetic-to-real displacement, we evaluate the effect of applying $\Delta_S^{\mathrm{s2r}}$ under three complementary conditions:
(i) on the source language $S$ from which it was computed,
(ii) on the intended target $T$, and
(iii) on the remaining languages $Q \neq S,T$, where neither the synthetic nor the real data played any role in constructing the vector.

The corresponding CER gains are shown in \cref{fig:improvements_type}.
As expected, improvements are largest when evaluating on the source $S$, confirming that each displacement retains a clear language-specific component tied to the domain from which it was estimated. However, despite this bias, the same vectors also yield improvements on the unrelated languages $Q$. Because $Q$ is never involved during analogy construction, this behavior reveals a transferable component of $\Delta_S^{\mathrm{s2r}}$ that is not tied to any particular linguistic distribution.

Together, these results show that each displacement combines two effects:
a source-dependent adjustment that reflects the specifics of its origin, and a broader correction that generalizes across languages and datasets. This dual structure explains why analogies remain effective even when source and target languages differ.



\subsection{Representation Quality}
To evaluate whether our analogy-based adaptation improves the internal representations learned by the models, beyond their zero-shot recognition accuracy, we conduct a \emph{linear probing} analysis~\cite{LP_alain2016}, which allows us to measure how linearly separable the relevant information is within the frozen feature space of a model. Concretely, if the model produces features that are well structured for the downstream task, then a simple linear classifier on top of those features should achieve strong performance with minimal training.

The results, summarized in \cref{fig:linear_probing}, show consistent gains across all architectures and datasets. Both single and multi-analogy variants outperform the baseline in CER and WER, demonstrating that analogy-based merging leads to feature spaces that are more compatible with real handwritten data. Notably, the multi-analogy configuration achieves the lowest error rates overall, indicating that aggregating multiple linguistic priors yields more robust and separable representations. The in-domain performance (dashed line) is included as an upper bound: analogy-based models consistently reduce the gap to this fully supervised reference.

Overall, this analysis reveals that the synthetic-to-real transformations estimated through analogy-based transfer also reshape the feature space in a way that improves linear separability under real-world handwriting distributions. In other words, analogy merging acts as a representation-learning mechanism that injects rich, language-informed structure into the model’s embeddings, even without any target-domain supervision.
\section{Conclusion}
We introduced a zero-shot synth-to-real adaptation method based on task analogies, that transfers synthetic-to-real corrections through task analogies. By transporting synth-to-real displacement vectors across languages, optionally merging several through linguistically informed weighting, we obtain significant zero-shot improvements over synthetic baselines on all evaluated architectures. The vectors capture both language-specific cues and a strong language-agnostic component, enabling gains even on languages not involved in analogy construction. Overall, our results show that adapting to real handwriting is itself a transferable operation, offering a scalable alternative to target-domain supervision.

\clearpage
{
    \small
    \bibliographystyle{ieeenat_fullname}
    \bibliography{main}
}

\let\addcontentsline\origaddcontentsline

\clearpage
\setcounter{page}{1}
\setcounter{table}{0}
\setcounter{figure}{0}

\maketitlesupplementary
\appendix

{
  \hypersetup{linkcolor=black}
  \tableofcontents
}
\vspace{.5cm}

\section{Datasets and splits}
In this section, we provide the specific details regarding the real handwritten datasets utilized in our study. We adhered strictly to the original train, validation, and test partitions defined for each dataset to ensure fair comparison and reproducibility against established benchmarks. Table \ref{tab:dataset-comparison} summarizes the key characteristics of these five datasets, detailing the language, historical context, number of writers, size of the vocabulary ($|\Sigma|$), and the number of samples in each partition used for our experiments. Some qualitative samples from these datasets are reported in~\Cref{fig:supp_datasets}.

\section{Synthetic data and Augmentations}

\subsection{Synthetic data}
\begin{figure*}
\centering
    \includegraphics[scale=0.7]{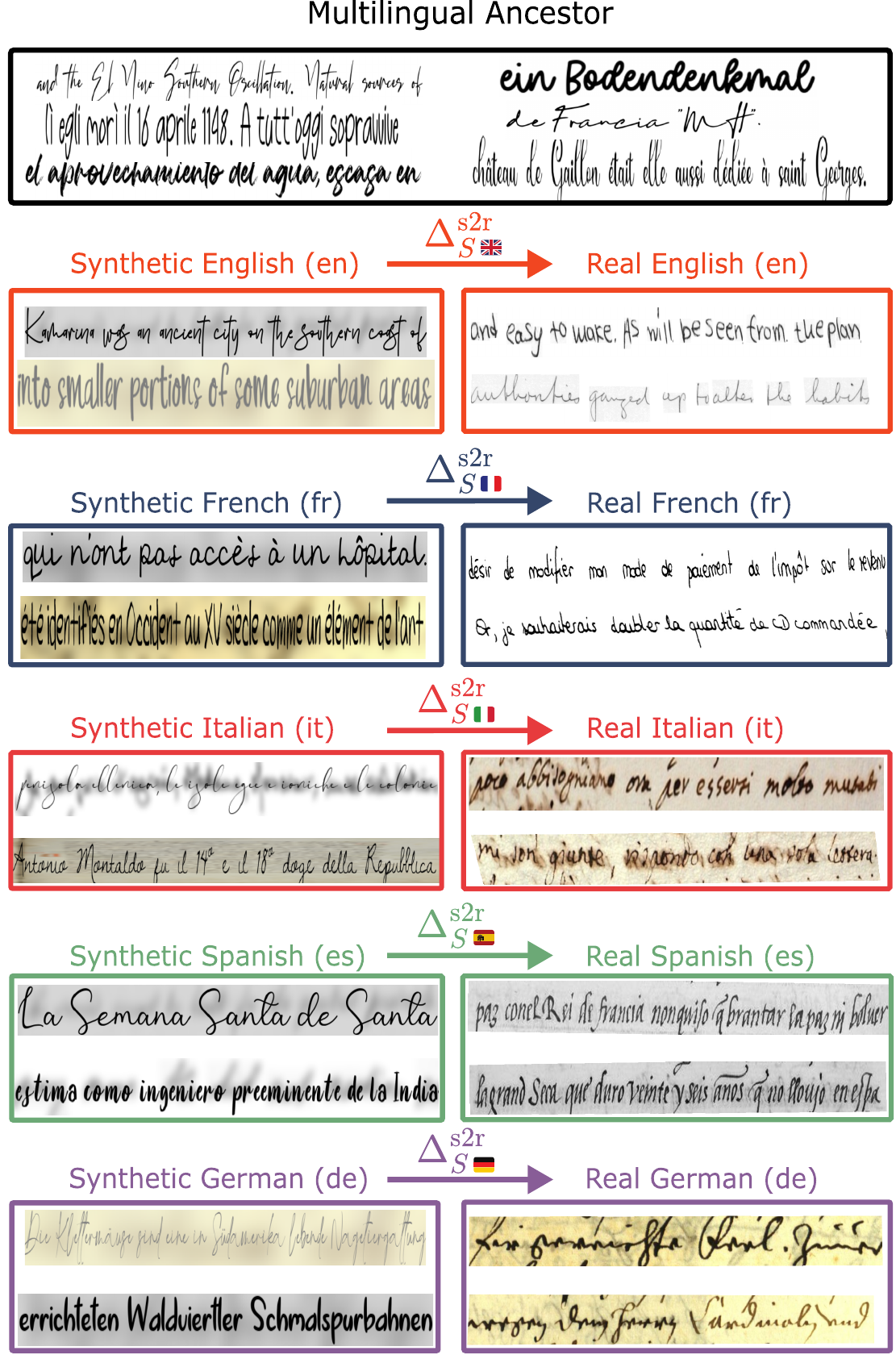}
    \caption{Samples from the synthetic datasets used to train the multilingual ancestor model and the synthetic child models for each language, alongside samples from the real datasets to perform the synthetic-to-real analogies.}
\label{fig:supp_datasets}
\end{figure*}

The foundation of our multilingual corpus utilizes approximately \num{3700} freely available handwritten fonts. Rendering was performed using the Python Imaging Library (PIL) and its \texttt{ImageDraw} class to ensure high-fidelity stroke preservation and pixel-level control. This process resulted in two distinct domains for task arithmetic:
\begin{itemize}
    \item \textbf{Ancestor Data:} Simplified, monochrome images with no augmentations or backgrounds, designed solely to capture generalized linguistic patterns (see~\Cref{fig:supp_datasets}).
    \item \textbf{Child Data:} Images rendered with complex backgrounds and textures, which serve as the basis for the aggressive geometric and morphological transformations detailed in~\Cref{ssec:augmentations} (see~\Cref{fig:supp_datasets}).
\end{itemize}
This controlled pipeline ensures a precise separation between the simplified ancestor parameter space and the complex, augmented child domain necessary for computing the synthetic-to-real transformation vector ($\Delta_{S}^{\mathrm{s2r}}$).

\subsection{Data augmentations}\label{ssec:augmentations}
We employed a data transformation pipeline that applies a sequence of augmentation techniques to simulate the visual noise, structural variance, and geometric distortions observed in real-world handwriting. These transformations are applied sequentially and are critical for mitigating the synthetic-to-real domain gap. Specifiacally, the following transformations are applied to the input image, each with a $50\%$ probability of activation:
\begin{enumerate}
    \item \textbf{Morphological Erosion:} The image undergoes erosion (using a $2 \times 2$ kernel for 1 iteration) to simulate slight loss of ink or thinning of strokes.
    \item \textbf{Random Affine Transformation:} A combination of geometric distortions is applied, including:
    \begin{itemize}
        \item Rotation: Applied within the range of $[-1, 1]$ degrees.
        \item Translation: Horizontal displacement up to $1\%$ and vertical displacement up to $5\%$ of the image size.
        \item Shear: Applied within the range of $[-1, 1]$ degrees on both the x and y axes.
    \end{itemize}
    \item \textbf{Random Perspective Distortion:} The image is subjected to a random perspective transformation with a distortion scale of $0.1$, simulating non-planar capture or slight document warping.
    \item \textbf{Random Rotation:} The image is rotated independently within a narrow range of $\pm 1$ degree.
\end{enumerate}
Crucially, this entire pipeline was applied uniformly to both the real HTR datasets and the highly augmented synthetic data used for training the child models. The simplified, monochrome data used to train the common ancestor model was explicitly excluded from these transformations, preserving its general, uncorrupted parameter space.

\begin{table}[t]
\centering
\caption{Handwritten document datasets considered, with an indication of the language, historical period, number of writers, number of samples in each partition, and size of the alphabet ($|\Sigma|$).}
\footnotesize
\renewcommand{\arraystretch}{1.2} 
\begin{tabular}{@{}p{2.0cm}p{0.25cm}c@{\hskip 0.35em}p{0.25cm}p{0.3cm}p{0.3cm}p{0.3cm}c@{}}
\toprule
\textbf{Dataset} & \textbf{Lang.} & \textbf{Period} & \textbf{No. writ} & \textbf{Train} & \textbf{Val} & \textbf{Test} & \textbf{$|\Sigma|$} \\
\midrule
IAM \cite{iamdatabase_marti_2002} & En & 1999 & 657 & 6.4k & 976 & 2.9k & 79 \\
Rimes \cite{rimes_2010} & Fr & 2011 & 1.3k & 10k & 1.1k & 778 & 100 \\
LAM \cite{cascianelli2022lam} & It & 17-18th c. & 1 & 20k & 2.5k & 3.5k & 89 \\
Rodrigo \cite{serrano-etal-2010-rodrigo} & Sp & 1545 & 1 & 20k & 1k & 5k & 115 \\
ICFHR$_{2016}$ \cite{icfhr_2016_sanchez} & De & 15-19th c. & \textit{unk.} & 8.2k & 1k & 1k & 91 \\
\bottomrule
\end{tabular}
\label{tab:dataset-comparison}
\end{table}

\begin{figure*}[t]
    \centering
    \setlength{\tabcolsep}{0pt}
    \begin{tabular}{ccc}
         \includegraphics[width=.34\linewidth]{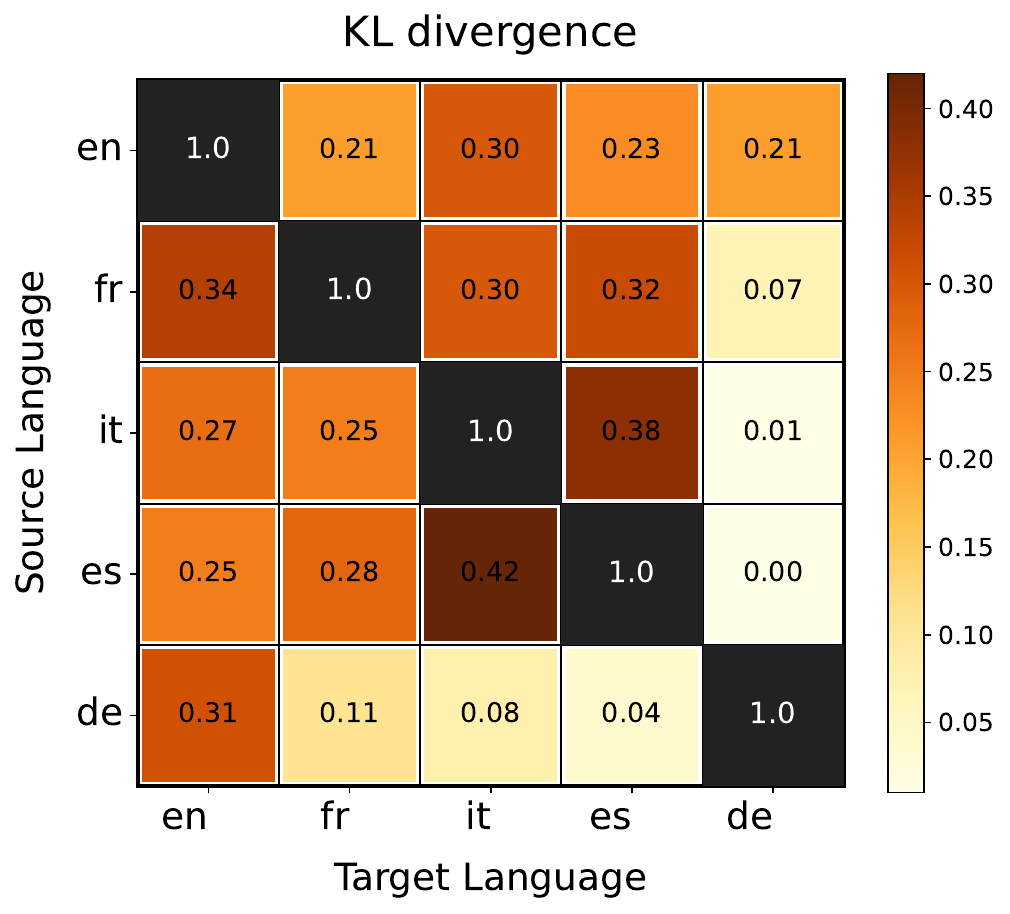} &  
         \includegraphics[width=.34\linewidth]{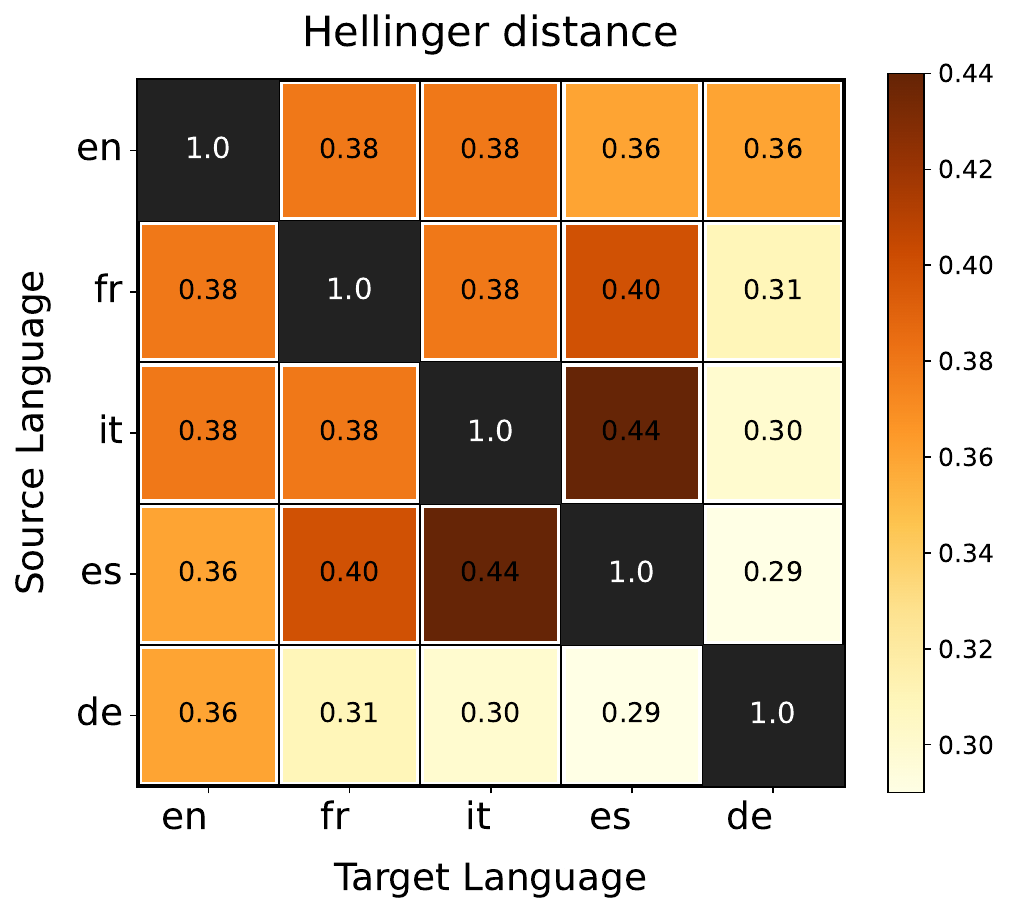} & 
         \includegraphics[width=.34\linewidth]{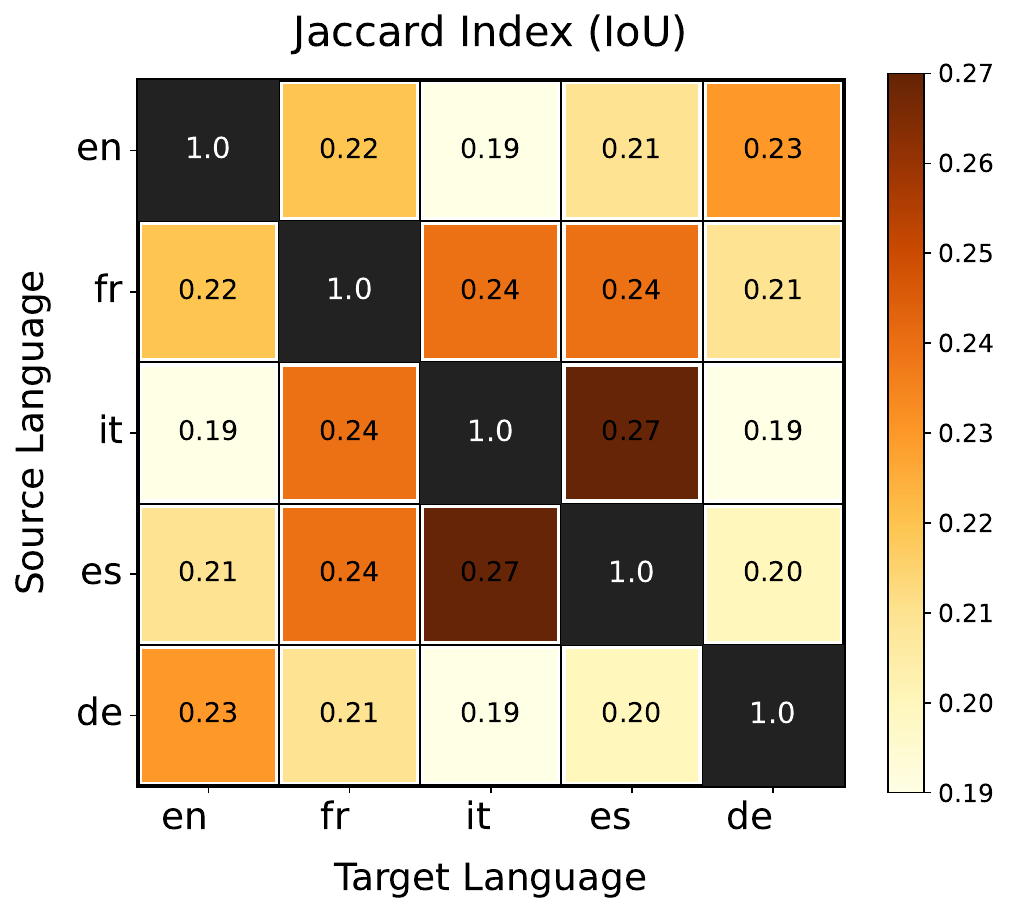}
    \end{tabular}
    \caption{Language similarity scores between each pair of considered Latin-alphabet languages.}
    \label{fig:lag_sim_scores}
\end{figure*}

\section{Linguistic scores}
The linguistic similarity scores: Kullback Leibler (KL) Divergence ($\beta_{\mathrm{KL}}$), Hellinger Distance ($\beta_{\mathrm{H}}$), and Jaccard Similarity ($\beta_{\mathrm{J}}$) are formally defined in the Methodology section (Section 2.3) of the main paper. Here, we provide the values of the scores for all language combinations used in the Multi Analogy merging strategy.
These scores were computed by using the $n$-gram frequency distributions derived exclusively from the validation splits of the WIT synthetic corpus \cite{wit_dataset}. We employed only synthetic text data for this analysis, ensuring that no linguistic information was ever extracted from the real HTR datasets. This adheres strictly to the zero-shot constraint of our framework. \Cref{fig:lag_sim_scores} reports the normalized $\beta$ scores, which represent the linguistic proximity used to weight the source analogy vectors ($\Delta_{S}^{\mathrm{s2r}}$).

\section{Architectural details}
This section provides an overview of the six HTR architectures employed in our experiments. Our selection encompasses a range of models, spanning from classical Convolutional-Recurrent Networks (CRNN/VAN) to modern Vision Transformer (ViT) based architectures (TrOCR/HTR-VT). Table \ref{tab:summary_architectures} summarizes the model characteristics, including alignment mechanism, total parameter count, and input resolution, highlighting the diversity of the models used to validate the robustness of our framework. 

For the CTC-based models (CRNN, VAN, and HTR-VT), we utilized a standard character tokenizer. The final vocabulary size for these models was defined by the number of unique characters observed across the multilingual synthetic corpus, plus the utility blank token for the CTC. For the Transformer-based TrOCR models, we adopted the originally used Byte-Pair Encoding (BPE) tokenizer \cite{radford2019language}. However, given that our experiments involved training all models \emph{from scratch} and were strictly focused on a closed set of five Latin-script languages (English, French, Italian, Spanish, and German), we adapted the vocabulary for computational efficiency and better domain focus.

\begin{table}[t]
\centering
\caption{Description of HTR architectures considered, including alignment type, vocabulary size, number of parameters, and input image sizes. \small FCN=Fully Convolutional Network; R=ResNet; T=Transformer; AR=Autoregressive.}

\footnotesize 
\begin{tabular}{@{}p{2.0cm}p{1.2cm}p{0.5cm}p{0.4cm}p{0.7cm}p{0.9cm}} 
\toprule
\textbf{Model} & \textbf{Type} & \textbf{Align.} & \textbf{Vocab} & \textbf{Params} & \textbf{Input size} \\
\midrule
\small CRNN~\cite{multidimensional_recurrent_puig_2017}   & CRNN         & CTC      & 95  & 9.6M  & 128$\times$1024 \\
\small VAN~\cite{endtoend_coquenet_2022}    & FCN          & CTC      & 95  & 2.7M  & 64$\times$1024 \\
\small HTR-VT~\cite{li_htr-vt_2025}       & R$_{18}$+ViT       & CTC      & 95 & 53.5M & 64$\times$512 \\
\small TrOCR(S)~\cite{trocr_li_2023}    & ViT+T.Dec  & AR  & 8192  & 33M   & 384$\times$384 \\
\small TrOCR(B)~\cite{trocr_li_2023}    & ViT+T.Dec  & AR  & 8192  & 299M  & 384$\times$384 \\
\small TrOCR(L)~\cite{trocr_li_2023}    & ViT+T.Dec  & AR  & 8192  & 523M  & 384$\times$384 \\
\bottomrule
\end{tabular}
\label{tab:summary_architectures}
\end{table}

\section{Experimental results}

\subsection{Main results by architecture}
In Tables~\ref{tab:supp_TrOCR_small}-\ref{tab:supp_HTR-VT}, we report the Character Error Rate (CER $\downarrow$) and Word Error Rate (WER $\downarrow$) across all real handwritten datasets for each of the six HTR architectures. Performance is compared against the \textbf{In-domain upper bound} and the \textbf{Baseline} ($\theta_T^{\text{syn}}$) zero-shot model. The results are grouped by single analogy and multiple analogy strategies, incorporating uniform and linguistically informed ($\beta_{KL}, \beta_{H}, \beta_{J}$) priors for zero-shot adaptation. 

\subsection{Linear probing results}
In~\Cref{tab:LP_all}, we report all the results (in terms of CER and WER) of the Linear Probing experiment evaluating the representation quality. We compare the Baseline against the best performing Linguistic Aware Single-Analogy and Multi-Analogy merged models, after fine-tuning only the final linear layer of the considered HTR architectures on the considered real datasets.

\clearpage

\begin{table*}[t]
    \caption{Performance across all real handwritten datasets for TrOCR (S). Values in parentheses are the improvement \wrt the Baseline.}\label{tab:supp_TrOCR_small}
    \centering
    \setlength{\tabcolsep}{.3em}
    \resizebox{\linewidth}{!}{%
    \footnotesize
    \begin{tabular}{l l ll l ll l ll l ll l ll l ll}
    \toprule
    && \multicolumn{2}{c}{\makecell{\textbf{IAM (en)}}} 
    && \multicolumn{2}{c}{\makecell{\textbf{Rimes (fr)}}}
    && \multicolumn{2}{c}{\makecell{\textbf{LAM (it)}}} 
    && \multicolumn{2}{c}{\makecell{\textbf{Rodrigo (es)}}} 
    && \multicolumn{2}{c}{\makecell{\textbf{$\text{ICFHR}_{2016}\text{ (de)}$}}} 
    && \multicolumn{2}{c}{\makecell{\textbf{AVG.}}} \\
    \cmidrule{3-4} \cmidrule{6-7} \cmidrule{9-10} \cmidrule{12-13} \cmidrule{15-16} \cmidrule{18-19}
    && \textbf{CER} $\downarrow$ & \textbf{WER} $\downarrow$ && \textbf{CER} $\downarrow$ & \textbf{WER} $\downarrow$ && \textbf{CER} $\downarrow$ & \textbf{WER} $\downarrow$ && \textbf{CER} $\downarrow$ & \textbf{WER} $\downarrow$ && \textbf{CER} $\downarrow$ & \textbf{WER} $\downarrow$ && \textbf{CER} $\downarrow$ & \textbf{WER} $\downarrow$ \\
    \midrule
    In-domain && 7.7 & 21.8 && 4.6 & 12.5 && 4.1 & 13.2 && 2.1 & 10.7 && 6.3 & 25.2 && 5.0 & 16.7 \\
    Baseline $\theta_T^{syn}$ && 16.4\deltazero & 39.7\deltazero && 24.1\deltazero & 59.1\deltazero && 46.9\deltazero & 78.9\deltazero && 30.3\deltazero & 76.6\deltazero && 80.7\deltazero & 107.8\deltazero && 39.7\deltazero & 72.4\deltazero \\
    \midrule
    \rowcolor{gray!20}
    \multicolumn{3}{l}{\textbf{Single analogy}} &&& & && & && & && & && & \\
    \midrule
    $\beta=1$  && 16.3\deltapos{-0.1} & 40.6\deltaneg{+0.9} && 29.0\deltaneg{+4.9} & 66.5\deltaneg{+7.4} && 36.4\deltapos{-10.5} & 72.0\deltapos{-6.9} && 23.8\deltapos{-6.5} & 70.3\deltapos{-6.3} && 84.0\deltaneg{+3.3} & 114.8\deltaneg{+7.0} && 37.9\deltapos{-1.8} & 72.8\deltaneg{+0.4} \\
    \midrule
    $\beta = \beta_{\text{KL}}$ && 12.4\deltapos{-4.0} & 33.4\deltapos{-6.3} && 24.8\deltaneg{+0.7} & 61.5\deltaneg{+2.4} && 33.8\deltapos{-13.1} & 68.2\deltapos{-10.7} && 23.8\deltapos{-6.5} & 70.3\deltapos{-6.3} && 80.3\deltapos{-0.4} & 107.6\deltapos{-0.2} && 35.0\deltapos{-4.7} & 68.2\deltapos{-4.2} \\
    $\beta = \beta_{\text{H}}$  && 12.1\deltapos{-4.3} & 32.9\deltapos{-6.8} && 25.9\deltaneg{+1.8} & 63.8\deltaneg{+4.7} && 33.7\deltapos{-13.2} & 68.4\deltapos{-10.5} && 24.1\deltapos{-6.2} & 70.5\deltapos{-6.1} && 80.9\deltaneg{+0.2} & 108.1\deltaneg{+0.3} && 35.3\deltapos{-4.4} & 68.7\deltapos{-3.7} \\
    $\beta = \beta_{\text{J}}$ && 15.9\deltapos{-0.5} & 40.0\deltaneg{+0.3} && 23.8\deltapos{-0.3} & 59.6\deltaneg{+0.5} && 33.6\deltapos{-13.3} & 68.4\deltapos{-10.5} && 23.9\deltapos{-6.4} & 70.4\deltapos{-6.2} && 80.3\deltapos{-0.4} & 107.7\deltapos{-0.1} && 35.5\deltapos{-4.2} & 69.2\deltapos{-3.2} \\
    \midrule
    \rowcolor{gray!20}
    \multicolumn{3}{l}{\textbf{Multiple analogy}} & && & && & && & && & && & \\
    \midrule
    $\beta=1$ && 20.5\deltaneg{+4.1} & 49.3\deltaneg{+9.6} && 28.3\deltaneg{+4.2} & 68.2\deltaneg{+9.1} && 29.8\deltapos{-17.1} & 66.7\deltapos{-12.2} && 27.7\deltapos{-2.6} & 71.5\deltapos{-5.1} && 167.7\deltaneg{+87.0} & 138.8\deltaneg{+31.0} && 54.8\deltaneg{+15.1} & 78.9\deltaneg{+6.5} \\
    $\beta=1/N$  && 15.6\deltapos{-0.8} & 40.3\deltaneg{+0.6} && 22.9\deltapos{-1.2} & 59.9\deltaneg{+0.8} && 28.7\deltapos{-18.2} & 62.7\deltapos{-16.2} && 24.6\deltapos{-5.7} & 68.7\deltapos{-7.9} && 81.2\deltaneg{+0.5} & 110.6\deltaneg{+2.8} && 34.6\deltapos{-5.1} & 68.4\deltapos{-4.0} \\
    \midrule
    $\beta = \beta_{\text{KL}}$ && 13.4\deltapos{-3.0} & 36.2\deltapos{-3.5} && 24.7\deltaneg{+0.6} & 63.5\deltaneg{+4.4} && 29.3\deltapos{-17.6} & 63.5\deltapos{-15.4} && 25.4\deltapos{-4.9} & 69.6\deltapos{-7.0} && 82.2\deltaneg{+1.5} & 108.6\deltaneg{+0.8} && 35.0\deltapos{-4.7} & 68.3\deltapos{-4.1} \\
    $\beta = \beta_{\text{H}}$ && 15.4\deltapos{-1.0} & 39.8\deltaneg{+0.1} && 22.5\deltapos{-1.6} & 59.1\deltapos{-.0} && 27.7\deltapos{-19.2} & 62.7\deltapos{-16.2} && 24.4\deltapos{-5.9} & 68.6\deltapos{-8.0} && 82.2\deltaneg{+1.5} & 111.1\deltaneg{+3.3} && 34.4\deltapos{-5.3} & 68.3\deltapos{-4.1} \\
    $\beta = \beta_{\text{J}}$ && 14.2\deltapos{-2.2} & 37.7\deltapos{-2.0} && 21.9\deltapos{-2.2} & 57.9\deltapos{-1.2} && 27.6\deltapos{-19.3} & 62.0\deltapos{-16.9} && 24.0\deltapos{-6.3} & 68.4\deltapos{-8.2} && 80.8\deltaneg{+0.1} & 109.7\deltaneg{+1.9} && 33.7\deltapos{-6.0} & 67.1\deltapos{-5.3} \\
   
    \bottomrule
    \end{tabular}}
\end{table*}
\begin{table*}[t]
    \caption{Performance across all real handwritten datasets for TrOCR (B). Values in parentheses are the improvement \wrt the Baseline.}\label{tab:supp_TrOCR_base}
    \centering
    \setlength{\tabcolsep}{.3em}
    \resizebox{\linewidth}{!}{%
    \footnotesize
    \begin{tabular}{l l ll l ll l ll l ll l ll l ll}
    \toprule
    && \multicolumn{2}{c}{\makecell{\textbf{IAM (en)}}} 
    && \multicolumn{2}{c}{\makecell{\textbf{Rimes (fr)}}}
    && \multicolumn{2}{c}{\makecell{\textbf{LAM (it)}}} 
    && \multicolumn{2}{c}{\makecell{\textbf{Rodrigo (es)}}} 
    && \multicolumn{2}{c}{\makecell{\textbf{$\text{ICFHR}_{2016}\text{ (de)}$}}} 
    && \multicolumn{2}{c}{\makecell{\textbf{AVG.}}} \\
    \cmidrule{3-4} \cmidrule{6-7} \cmidrule{9-10} \cmidrule{12-13} \cmidrule{15-16} \cmidrule{18-19}
    && \textbf{CER} $\downarrow$ & \textbf{WER} $\downarrow$ && \textbf{CER} $\downarrow$ & \textbf{WER} $\downarrow$ && \textbf{CER} $\downarrow$ & \textbf{WER} $\downarrow$ && \textbf{CER} $\downarrow$ & \textbf{WER} $\downarrow$ && \textbf{CER} $\downarrow$ & \textbf{WER} $\downarrow$ && \textbf{CER} $\downarrow$ & \textbf{WER} $\downarrow$ \\
    \midrule
    In-domain && 6.3 & 17.7 && 3.8 & 9.6 && 3.5 & 11.0 && 2.2 & 10.7 && 6.4 & 24.8 && 4.4 & 14.8 \\
    Baseline $\theta_T^{syn}$ && 14.8\deltazero & 37.7\deltazero && 22.2\deltazero & 51.8\deltazero && 40.3\deltazero & 75.8\deltazero && 31.6\deltazero & 75.7\deltazero && 79.6\deltazero & 104.3\deltazero && 37.7\deltazero & 69.1\deltazero \\
    \midrule
    \rowcolor{gray!20}
    \multicolumn{3}{l}{\textbf{Single analogy}} &&& & && & && & && & && & \\
    \midrule
    $\beta=1$  && 19.5\deltaneg{+4.7} & 45.6\deltaneg{+7.9} && 27.1\deltaneg{+4.9} & 61.6\deltaneg{+9.8} && 32.4\deltapos{-7.9} & 67.4\deltapos{-8.4} && 26.5\deltapos{-5.1} & 72.4\deltapos{-3.3} && 79.0\deltapos{-0.6} & 107.5\deltaneg{+3.2} && 36.9\deltapos{-0.8} & 70.9\deltaneg{+1.8} \\
    \midrule
    $\beta = \beta_{\text{KL}}$ && 11.5\deltapos{-3.3} & 31.8\deltapos{-5.9} && 19.2\deltapos{-3.0} & 49.8\deltapos{-2.0} && 32.2\deltapos{-8.1} & 69.2\deltapos{-6.6} && 22.5\deltapos{-9.1} & 67.2\deltapos{-8.5} && 79.2\deltapos{-0.4} & 105.1\deltaneg{+0.8} && 32.9\deltapos{-4.8} & 64.6\deltapos{-4.5} \\
    $\beta = \beta_{\text{H}}$  && 11.4\deltapos{-3.4} & 31.4\deltapos{-6.3} && 20.0\deltapos{-2.2} & 51.7\deltapos{-0.1} && 32.0\deltapos{-8.3} & 68.4\deltapos{-7.4} && 22.6\deltapos{-9.0} & 67.2\deltapos{-8.5} && 79.2\deltapos{-0.4} & 105.6\deltaneg{+1.3} && 33.0\deltapos{-4.7} & 64.9\deltapos{-4.2} \\
    $\beta = \beta_{\text{J}}$ && 13.7\deltapos{-1.1} & 35.3\deltapos{-2.4} && 19.1\deltapos{-3.1} & 49.3\deltapos{-2.5} && 32.3\deltapos{-8.0} & 68.6\deltapos{-7.2} && 24.2\deltapos{-7.4} & 68.1\deltapos{-7.6} && 79.0\deltapos{-0.6} & 105.4\deltaneg{+1.1} && 33.7\deltapos{-4.0} & 65.3\deltapos{-3.8} \\
    \midrule
    \rowcolor{gray!20}
    \multicolumn{3}{l}{\textbf{Multiple analogy}} & && & && & && & && & && & \\
    \midrule
    $\beta=1$ && 17.9\deltaneg{+3.1} & 43.9\deltaneg{+6.2} && 19.0\deltapos{-3.2} & 50.9\deltapos{-0.9} && 27.3\deltapos{-13.0} & 61.8\deltapos{-14.0} && 22.4\deltapos{-9.2} & 65.0\deltapos{-10.7} && 82.6\deltaneg{+3.0} & 114.1\deltaneg{+9.8} && 33.8\deltapos{-3.9} & 67.1\deltapos{-2.0} \\
    $\beta=1/N$  && 13.8\deltapos{-1.0} & 36.2\deltapos{-1.5} && 18.9\deltapos{-3.3} & 50.8\deltapos{-1.0} && 26.0\deltapos{-14.3} & 59.9\deltapos{-15.9} && 21.6\deltapos{-10.0} & 64.8\deltapos{-10.9} && 78.9\deltapos{-0.7} & 109.1\deltaneg{+4.8} && 31.8\deltapos{-5.9} & 64.2\deltapos{-4.9} \\
    \midrule
    $\beta = \beta_{\text{KL}}$ && 15.4\deltaneg{+0.6} & 39.5\deltaneg{+1.8} && 17.2\deltapos{-5.0} & 47.1\deltapos{-4.7} && 27.8\deltapos{-12.5} & 62.7\deltapos{-13.1} && 22.4\deltapos{-9.2} & 64.8\deltapos{-10.9} && 80.0\deltaneg{+0.4} & 105.4\deltaneg{+1.1} && 32.6\deltapos{-5.1} & 63.9\deltapos{-5.2} \\
    $\beta = \beta_{\text{H}}$ && 13.6\deltapos{-1.2} & 35.9\deltapos{-1.8} && 19.8\deltapos{-2.4} & 52.3\deltaneg{+0.5} && 26.1\deltapos{-14.2} & 60.2\deltapos{-15.6} && 21.5\deltapos{-10.1} & 64.4\deltapos{-11.3} && 81.4\deltaneg{+1.8} & 113.3\deltaneg{+9.0} && 32.5\deltapos{-5.2} & 65.2\deltapos{-3.9} \\
    $\beta = \beta_{\text{J}}$ && 15.1\deltaneg{+0.3} & 38.8\deltaneg{+1.1} && 18.1\deltapos{-4.1} & 48.7\deltapos{-3.1} && 26.3\deltapos{-14.0} & 60.7\deltapos{-15.1} && 21.8\deltapos{-9.8} & 64.4\deltapos{-11.3} && 79.3\deltapos{-0.3} & 109.6\deltaneg{+5.3} && 32.1\deltapos{-5.6} & 64.4\deltapos{-4.7} \\
   
    \bottomrule
    \end{tabular}}
\end{table*}
\begin{table*}[t]
    \caption{Performance across all real handwritten datasets for TrOCR (L). Values in parentheses are the improvement \wrt the Baseline.}\label{tab:supp_TrOCR_large}
    \centering
    \setlength{\tabcolsep}{.3em}
    \resizebox{\linewidth}{!}{%
    \footnotesize
    \begin{tabular}{l l ll l ll l ll l ll l ll l ll}
    \toprule
    && \multicolumn{2}{c}{\makecell{\textbf{IAM (en)}}} 
    && \multicolumn{2}{c}{\makecell{\textbf{Rimes (fr)}}}
    && \multicolumn{2}{c}{\makecell{\textbf{LAM (it)}}} 
    && \multicolumn{2}{c}{\makecell{\textbf{Rodrigo (es)}}} 
    && \multicolumn{2}{c}{\makecell{\textbf{$\text{ICFHR}_{2016}\text{ (de)}$}}} 
    && \multicolumn{2}{c}{\makecell{\textbf{AVG.}}} \\
    \cmidrule{3-4} \cmidrule{6-7} \cmidrule{9-10} \cmidrule{12-13} \cmidrule{15-16} \cmidrule{18-19}
    && \textbf{CER} $\downarrow$ & \textbf{WER} $\downarrow$ && \textbf{CER} $\downarrow$ & \textbf{WER} $\downarrow$ && \textbf{CER} $\downarrow$ & \textbf{WER} $\downarrow$ && \textbf{CER} $\downarrow$ & \textbf{WER} $\downarrow$ && \textbf{CER} $\downarrow$ & \textbf{WER} $\downarrow$ && \textbf{CER} $\downarrow$ & \textbf{WER} $\downarrow$ \\
    \midrule
    In-domain && 6.7 & 19.3 && 3.9 & 9.6 && 3.7 & 11.9 && 1.8 & 9.0 && 5.2 & 20.5 && 4.3 & 14.1 \\
    Baseline $\theta_T^{syn}$ && 14.0\deltazero & 35.0\deltazero && 19.1\deltazero & 48.3\deltazero && 40.2\deltazero & 73.7\deltazero && 30.2\deltazero & 76.3\deltazero && 77.3\deltazero & 105.7\deltazero && 36.2\deltazero & 67.8\deltazero \\
    \midrule
    \rowcolor{gray!20}
    \multicolumn{3}{l}{\textbf{Single analogy}} &&& & && & && & && & && & \\
    \midrule
    $\beta=1$  && 17.0\deltaneg{+3.0} & 42.7\deltaneg{+7.7} && 26.1\deltaneg{+7.0} & 63.0\deltaneg{+14.7} && 33.6\deltapos{-6.6} & 71.7\deltapos{-2.0} && 26.3\deltapos{-3.9} & 76.0\deltapos{-0.3} && 73.6\deltapos{-3.7} & 112.1\deltaneg{+6.4} && 35.3\deltapos{-0.9} & 73.1\deltaneg{+5.3} \\
    \midrule
    $\beta = \beta_{\text{KL}}$ && 12.2\deltapos{-1.8} & 33.3\deltapos{-1.7} && 20.0\deltaneg{+0.9} & 53.6\deltaneg{+5.3} && 23.6\deltapos{-16.6} & 56.5\deltapos{-17.2} && 27.3\deltapos{-2.9} & 72.8\deltapos{-3.5} && 79.1\deltaneg{+1.8} & 106.4\deltaneg{+0.7} && 32.4\deltapos{-3.8} & 64.5\deltapos{-3.3} \\
    $\beta = \beta_{\text{H}}$  && 14.4\deltaneg{+0.4} & 38.0\deltaneg{+3.0} && 15.6\deltapos{-3.5} & 44.3\deltapos{-4.0} && 23.0\deltapos{-17.2} & 55.8\deltapos{-17.9} && 26.0\deltapos{-4.2} & 71.5\deltapos{-4.8} && 75.4\deltapos{-1.9} & 108.3\deltaneg{+2.6} && 30.9\deltapos{-5.3} & 63.6\deltapos{-4.2} \\
    $\beta = \beta_{\text{J}}$ && 12.9\deltapos{-1.1} & 34.9\deltapos{-0.1} && 17.9\deltapos{-1.2} & 49.6\deltaneg{+1.3} && 22.4\deltapos{-17.8} & 54.5\deltapos{-19.2} && 26.1\deltapos{-4.1} & 71.5\deltapos{-4.8} && 77.1\deltapos{-0.2} & 112.7\deltaneg{+7.0} && 31.3\deltapos{-4.9} & 64.6\deltapos{-3.2} \\
    \midrule
    \rowcolor{gray!20}
    \multicolumn{3}{l}{\textbf{Multiple analogy}} & && & && & && & && & && & \\
    \midrule
    $\beta=1$ && 20.2\deltaneg{+6.2} & 48.5\deltaneg{+13.5} && 19.5\deltaneg{+0.4} & 53.2\deltaneg{+4.9} && 23.8\deltapos{-16.4} & 57.2\deltapos{-16.5} && 48.1\deltaneg{+17.9} & 88.3\deltaneg{+12.0} && 84.0\deltaneg{+6.7} & 122.3\deltaneg{+16.6} && 39.1\deltaneg{+2.9} & 73.9\deltaneg{+6.1} \\
    $\beta=1/N$  && 14.7\deltaneg{+0.7} & 38.5\deltaneg{+3.5} && 16.1\deltapos{-3.0} & 45.2\deltapos{-3.1} && 22.5\deltapos{-17.7} & 55.0\deltapos{-18.7} && 26.2\deltapos{-4.0} & 71.5\deltapos{-4.8} && 78.0\deltaneg{+0.7} & 114.5\deltaneg{+8.8} && 31.5\deltapos{-4.7} & 64.9\deltapos{-2.9} \\
    \midrule
    $\beta = \beta_{\text{KL}}$ && 12.2\deltapos{-1.8} & 33.3\deltapos{-1.7} && 20.0\deltaneg{+0.9} & 53.6\deltaneg{+5.3} && 23.6\deltapos{-16.6} & 56.5\deltapos{-17.2} && 27.3\deltapos{-2.9} & 72.8\deltapos{-3.5} && 78.9\deltaneg{+1.6} & 107.4\deltaneg{+1.7} && 32.4\deltapos{-3.8} & 64.7\deltapos{-3.1} \\
    $\beta = \beta_{\text{H}}$ && 14.4\deltaneg{+0.4} & 38.0\deltaneg{+3.0} && 15.6\deltapos{-3.5} & 44.3\deltapos{-4.0} && 23.0\deltapos{-17.2} & 55.8\deltapos{-17.9} && 26.0\deltapos{-4.2} & 71.5\deltapos{-4.8} && 75.4\deltapos{-1.9} & 108.3\deltaneg{+2.6} && 30.9\deltapos{-5.3} & 63.6\deltapos{-4.2} \\
    $\beta = \beta_{\text{J}}$ && 12.9\deltapos{-1.1} & 34.9\deltapos{-0.1} && 17.9\deltapos{-1.2} & 49.6\deltaneg{+1.3} && 22.4\deltapos{-17.8} & 54.5\deltapos{-19.2} && 26.1\deltapos{-4.1} & 71.5\deltapos{-4.8} && 77.1\deltapos{-0.2} & 112.7\deltaneg{+7.0} && 31.3\deltapos{-4.9} & 64.6\deltapos{-3.2} \\
   
    \bottomrule
    \end{tabular}}
\end{table*}
\begin{table*}[t]
    \caption{Performance across all real handwritten datasets for CRNN. Values in parentheses are the improvement \wrt the Baseline.}\label{tab:supp_CRNN}
    \centering
    \setlength{\tabcolsep}{.3em}
    \resizebox{\linewidth}{!}{%
    \footnotesize
    \begin{tabular}{l l ll l ll l ll l ll l ll l ll}
    \toprule
    && \multicolumn{2}{c}{\makecell{\textbf{IAM (en)}}} 
    && \multicolumn{2}{c}{\makecell{\textbf{Rimes (fr)}}}
    && \multicolumn{2}{c}{\makecell{\textbf{LAM (it)}}} 
    && \multicolumn{2}{c}{\makecell{\textbf{Rodrigo (es)}}} 
    && \multicolumn{2}{c}{\makecell{\textbf{$\text{ICFHR}_{2016}\text{ (de)}$}}} 
    && \multicolumn{2}{c}{\makecell{\textbf{AVG.}}} \\
    \cmidrule{3-4} \cmidrule{6-7} \cmidrule{9-10} \cmidrule{12-13} \cmidrule{15-16} \cmidrule{18-19}
    && \textbf{CER} $\downarrow$ & \textbf{WER} $\downarrow$ && \textbf{CER} $\downarrow$ & \textbf{WER} $\downarrow$ && \textbf{CER} $\downarrow$ & \textbf{WER} $\downarrow$ && \textbf{CER} $\downarrow$ & \textbf{WER} $\downarrow$ && \textbf{CER} $\downarrow$ & \textbf{WER} $\downarrow$ && \textbf{CER} $\downarrow$ & \textbf{WER} $\downarrow$ \\
    \midrule
    In-domain && 6.2 & 22.0 && 4.1 & 13.8 && 3.5 & 12.5 && 1.0 & 5.5 && 6.3 & 26.1 && 4.2 & 16.0 \\
    Baseline $\theta_T^{syn}$ && 17.0\deltazero & 48.4\deltazero && 23.0\deltazero & 63.3\deltazero && 37.3\deltazero & 78.2\deltazero && 32.9\deltazero & 87.0\deltazero && 83.8\deltazero & 100.9\deltazero && 38.8\deltazero & 75.6\deltazero \\
    \midrule
    \rowcolor{gray!20}
    \multicolumn{3}{l}{\textbf{Single analogy}} &&& & && & && & && & && & \\
    \midrule
    $\beta=1$  && 18.3\deltaneg{+1.3} & 50.4\deltaneg{+2.0} && 23.0\deltapos{-.0} & 63.3\deltapos{-.0} && 38.2\deltaneg{+0.9} & 73.7\deltapos{-4.5} && 21.3\deltapos{-11.6} & 76.7\deltapos{-10.3} && 78.8\deltapos{-5.0} & 102.5\deltaneg{+1.6} && 35.9\deltapos{-2.9} & 73.3\deltapos{-2.3} \\
    \midrule
    $\beta = \beta_{\text{KL}}$ && 15.3\deltapos{-1.7} & 44.7\deltapos{-3.7} && 23.0\deltapos{-.0} & 64.3\deltaneg{+1.0} && 32.3\deltapos{-5.0} & 71.7\deltapos{-6.5} && 21.7\deltapos{-11.2} & 77.2\deltapos{-9.8} && 84.5\deltaneg{+0.7} & 100.8\deltapos{-0.1} && 35.4\deltapos{-3.4} & 71.7\deltapos{-3.9} \\
    $\beta = \beta_{\text{H}}$  && 14.4\deltapos{-2.6} & 42.7\deltapos{-5.7} && 22.8\deltapos{-0.2} & 63.8\deltaneg{+0.5} && 32.3\deltapos{-5.0} & 71.7\deltapos{-6.5} && 21.5\deltapos{-11.4} & 76.9\deltapos{-10.1} && 85.5\deltaneg{+1.7} & 100.9\deltapos{-.0} && 35.3\deltapos{-3.5} & 71.2\deltapos{-4.4} \\
    $\beta = \beta_{\text{J}}$ && 16.8\deltapos{-0.2} & 47.3\deltapos{-1.1} && 22.8\deltapos{-0.2} & 63.9\deltaneg{+0.6} && 32.3\deltapos{-5.0} & 71.7\deltapos{-6.5} && 23.0\deltapos{-9.9} & 78.4\deltapos{-8.6} && 84.5\deltaneg{+0.7} & 100.9\deltapos{-.0} && 35.9\deltapos{-2.9} & 72.4\deltapos{-3.2} \\
    \midrule
    \rowcolor{gray!20}
    \multicolumn{3}{l}{\textbf{Multiple analogy}} & && & && & && & && & && & \\
    \midrule
    $\beta=1$ && 15.8\deltapos{-1.2} & 45.5\deltapos{-2.9} && 20.0\deltapos{-3.0} & 58.7\deltapos{-4.6} && 100.0\deltaneg{+62.7} & 100.0\deltaneg{+21.8} && 21.2\deltapos{-11.7} & 75.9\deltapos{-11.1} && 100.0\deltaneg{+16.2} & 100.0\deltapos{-0.9} && 51.4\deltaneg{+12.6} & 76.0\deltaneg{+0.4} \\
    $\beta=1/N$  && 15.8\deltapos{-1.2} & 45.5\deltapos{-2.9} && 20.1\deltapos{-2.9} & 58.7\deltapos{-4.6} && 32.5\deltapos{-4.8} & 71.1\deltapos{-7.1} && 21.2\deltapos{-11.7} & 75.9\deltapos{-11.1} && 84.4\deltaneg{+0.6} & 101.6\deltaneg{+0.7} && 34.8\deltapos{-4.0} & 70.6\deltapos{-5.0} \\
    \midrule
    $\beta = \beta_{\text{KL}}$ && 15.3\deltapos{-1.7} & 44.6\deltapos{-3.8} && 20.1\deltapos{-2.9} & 58.6\deltapos{-4.7} && 33.6\deltapos{-3.7} & 73.2\deltapos{-5.0} && 21.6\deltapos{-11.3} & 76.5\deltapos{-10.5} && 84.3\deltaneg{+0.5} & 100.9\deltapos{-.0} && 35.0\deltapos{-3.8} & 70.8\deltapos{-4.8} \\
    $\beta = \beta_{\text{H}}$ && 15.9\deltapos{-1.1} & 45.8\deltapos{-2.6} && 20.0\deltapos{-3.0} & 58.9\deltapos{-4.4} && 33.4\deltapos{-3.9} & 72.3\deltapos{-5.9} && 20.9\deltapos{-12.0} & 75.4\deltapos{-11.6} && 85.1\deltaneg{+1.3} & 101.0\deltaneg{+0.1} && 35.1\deltapos{-3.7} & 70.7\deltapos{-4.9} \\
    $\beta = \beta_{\text{J}}$ && 15.5\deltapos{-1.5} & 45.0\deltapos{-3.4} && 20.0\deltapos{-3.0} & 58.8\deltapos{-4.5} && 32.3\deltapos{-5.0} & 71.0\deltapos{-7.2} && 21.4\deltapos{-11.5} & 76.4\deltapos{-10.6} && 84.3\deltaneg{+0.5} & 101.3\deltaneg{+0.4} && 34.7\deltapos{-4.1} & 70.5\deltapos{-5.1} \\
   
    \bottomrule
    \end{tabular}}
\end{table*}
\begin{table*}[t]
    \caption{Performance across all real handwritten datasets for VAN. Values in parentheses are the improvement \wrt the Baseline.}\label{tab:supp_VAN}
    \centering
    \setlength{\tabcolsep}{.3em}
    \resizebox{\linewidth}{!}{%
    \footnotesize
    \begin{tabular}{l l ll l ll l ll l ll l ll l ll}
    \toprule
    && \multicolumn{2}{c}{\makecell{\textbf{IAM (en)}}} 
    && \multicolumn{2}{c}{\makecell{\textbf{Rimes (fr)}}}
    && \multicolumn{2}{c}{\makecell{\textbf{LAM (it)}}} 
    && \multicolumn{2}{c}{\makecell{\textbf{Rodrigo (es)}}} 
    && \multicolumn{2}{c}{\makecell{\textbf{$\text{ICFHR}_{2016}\text{ (de)}$}}} 
    && \multicolumn{2}{c}{\makecell{\textbf{AVG.}}} \\
    \cmidrule{3-4} \cmidrule{6-7} \cmidrule{9-10} \cmidrule{12-13} \cmidrule{15-16} \cmidrule{18-19}
    && \textbf{CER} $\downarrow$ & \textbf{WER} $\downarrow$ && \textbf{CER} $\downarrow$ & \textbf{WER} $\downarrow$ && \textbf{CER} $\downarrow$ & \textbf{WER} $\downarrow$ && \textbf{CER} $\downarrow$ & \textbf{WER} $\downarrow$ && \textbf{CER} $\downarrow$ & \textbf{WER} $\downarrow$ && \textbf{CER} $\downarrow$ & \textbf{WER} $\downarrow$ \\
    \midrule
    In-domain && 6.7 & 25.5 && 5.5 & 20.4 && 3.7 & 13.5 && 1.7 & 11.2 && 9.4 & 38.5 && 5.4 & 21.8 \\
    Baseline $\theta_T^{syn}$ && 17.6\deltazero & 52.6\deltazero && 19.9\deltazero & 62.9\deltazero && 33.6\deltazero & 79.3\deltazero && 26.8\deltazero & 84.5\deltazero && 78.2\deltazero & 102.0\deltazero && 35.2\deltazero & 76.3\deltazero \\
    \midrule
    \rowcolor{gray!20}
    \multicolumn{3}{l}{\textbf{Single analogy}} &&& & && & && & && & && & \\
    \midrule
    $\beta=1$  && 22.6\deltaneg{+5.0} & 62.6\deltaneg{+10.0} && 27.2\deltaneg{+7.3} & 74.1\deltaneg{+11.2} && 28.7\deltapos{-4.9} & 74.0\deltapos{-5.3} && 20.5\deltapos{-6.3} & 75.1\deltapos{-9.4} && 73.4\deltapos{-4.8} & 104.3\deltaneg{+2.3} && 34.5\deltapos{-0.7} & 78.0\deltaneg{+1.7} \\
    \midrule
    $\beta = \beta_{\text{KL}}$ && 14.1\deltapos{-3.5} & 45.2\deltapos{-7.4} && 20.5\deltaneg{+0.6} & 65.0\deltaneg{+2.1} && 27.6\deltapos{-6.0} & 73.3\deltapos{-6.0} && 19.9\deltapos{-6.9} & 75.4\deltapos{-9.1} && 76.3\deltapos{-1.9} & 101.6\deltapos{-0.4} && 31.7\deltapos{-3.5} & 72.1\deltapos{-4.2} \\
    $\beta = \beta_{\text{H}}$  && 13.9\deltapos{-3.7} & 44.8\deltapos{-7.8} && 21.4\deltaneg{+1.5} & 66.8\deltaneg{+3.9} && 27.5\deltapos{-6.1} & 73.4\deltapos{-5.9} && 19.7\deltapos{-7.1} & 75.2\deltapos{-9.3} && 75.8\deltapos{-2.4} & 101.3\deltapos{-0.7} && 31.7\deltapos{-3.5} & 72.3\deltapos{-4.0} \\
    $\beta = \beta_{\text{J}}$ && 17.1\deltapos{-0.5} & 51.9\deltapos{-0.7} && 20.2\deltaneg{+0.3} & 64.3\deltaneg{+1.4} && 28.7\deltapos{-4.9} & 74.5\deltapos{-4.8} && 20.8\deltapos{-6.0} & 76.6\deltapos{-7.9} && 76.3\deltapos{-1.9} & 101.6\deltapos{-0.4} && 32.6\deltapos{-2.6} & 73.8\deltapos{-2.5} \\
    \midrule
    \rowcolor{gray!20}
    \multicolumn{3}{l}{\textbf{Multiple analogy}} & && & && & && & && & && & \\
    \midrule
    $\beta=1$ && 17.4\deltapos{-0.2} & 53.3\deltaneg{+0.7} && 19.4\deltapos{-0.5} & 60.9\deltapos{-2.0} && 21.8\deltapos{-11.8} & 62.9\deltapos{-16.4} && 18.5\deltapos{-8.3} & 73.1\deltapos{-11.4} && 74.0\deltapos{-4.2} & 101.2\deltapos{-0.8} && 30.2\deltapos{-5.0} & 70.3\deltapos{-6.0} \\
    $\beta=1/N$  && 17.4\deltapos{-0.2} & 53.3\deltaneg{+0.7} && 19.4\deltapos{-0.5} & 60.9\deltapos{-2.0} && 21.8\deltapos{-11.8} & 62.9\deltapos{-16.4} && 18.5\deltapos{-8.3} & 73.1\deltapos{-11.4} && 74.0\deltapos{-4.2} & 101.2\deltapos{-0.8} && 30.2\deltapos{-5.0} & 70.3\deltapos{-6.0} \\
    \midrule
    $\beta = \beta_{\text{KL}}$ && 18.1\deltaneg{+0.5} & 55.3\deltaneg{+2.7} && 18.5\deltapos{-1.4} & 59.2\deltapos{-3.7} && 22.6\deltapos{-11.0} & 64.1\deltapos{-15.2} && 19.7\deltapos{-7.1} & 75.7\deltapos{-8.8} && 76.3\deltapos{-1.9} & 101.4\deltapos{-0.6} && 31.0\deltapos{-4.2} & 71.1\deltapos{-5.2} \\
    $\beta = \beta_{\text{H}}$ && 18.1\deltaneg{+0.5} & 55.3\deltaneg{+2.7} && 18.5\deltapos{-1.4} & 59.2\deltapos{-3.7} && 22.6\deltapos{-11.0} & 64.1\deltapos{-15.2} && 19.7\deltapos{-7.1} & 75.7\deltapos{-8.8} && 76.3\deltapos{-1.9} & 101.4\deltapos{-0.6} && 31.0\deltapos{-4.2} & 71.1\deltapos{-5.2} \\
    $\beta = \beta_{\text{J}}$ && 16.3\deltapos{-1.3} & 51.1\deltapos{-1.5} && 18.9\deltapos{-1.0} & 59.9\deltapos{-3.0} && 21.9\deltapos{-11.7} & 63.4\deltapos{-15.9} && 18.5\deltapos{-8.3} & 73.3\deltapos{-11.2} && 73.8\deltapos{-4.4} & 101.3\deltapos{-0.7} && 29.9\deltapos{-5.3} & 69.8\deltapos{-6.5} \\
   
    \bottomrule
    \end{tabular}}
\end{table*}
\begin{table*}[t]
    \caption{Performance across all real handwritten datasets for HTR-VIT. Values in parentheses are the improvement \wrt the Baseline.}\label{tab:supp_HTR-VT}
    \centering
    \setlength{\tabcolsep}{.3em}
    \resizebox{\linewidth}{!}{%
    \footnotesize
    \begin{tabular}{l l ll l ll l ll l ll l ll l ll}
    \toprule
    && \multicolumn{2}{c}{\makecell{\textbf{IAM (en)}}} 
    && \multicolumn{2}{c}{\makecell{\textbf{Rimes (fr)}}}
    && \multicolumn{2}{c}{\makecell{\textbf{LAM (it)}}} 
    && \multicolumn{2}{c}{\makecell{\textbf{Rodrigo (es)}}} 
    && \multicolumn{2}{c}{\makecell{\textbf{$\text{ICFHR}_{2016}\text{ (de)}$}}} 
    && \multicolumn{2}{c}{\makecell{\textbf{AVG.}}} \\
    \cmidrule{3-4} \cmidrule{6-7} \cmidrule{9-10} \cmidrule{12-13} \cmidrule{15-16} \cmidrule{18-19}
    && \textbf{CER} $\downarrow$ & \textbf{WER} $\downarrow$ && \textbf{CER} $\downarrow$ & \textbf{WER} $\downarrow$ && \textbf{CER} $\downarrow$ & \textbf{WER} $\downarrow$ && \textbf{CER} $\downarrow$ & \textbf{WER} $\downarrow$ && \textbf{CER} $\downarrow$ & \textbf{WER} $\downarrow$ && \textbf{CER} $\downarrow$ & \textbf{WER} $\downarrow$ \\
    \midrule
    In-domain && 7.9 & 29.6 && 5.5 & 20.4 && 3.6 & 13.3 && 2.0 & 12.9 && 7.5 & 33.0 && 5.3 & 21.8 \\
    Baseline $\theta_T^{syn}$ && 21.9\deltazero & 62.3\deltazero && 31.2\deltazero & 77.2\deltazero && 45.4\deltazero & 89.5\deltazero && 26.2\deltazero & 79.4\deltazero && 84.0\deltazero & 100.4\deltazero && 41.7\deltazero & 81.8\deltazero \\
    \midrule
    \rowcolor{gray!20}
    \multicolumn{3}{l}{\textbf{Single analogy}} &&& & && & && & && & && & \\
    \midrule
    $\beta=1$   && 28.4\deltaneg{+6.5} & 73.0\deltaneg{+10.7} && 38.3\deltaneg{+7.1} & 86.9\deltaneg{+9.7} && 34.4\deltapos{-11.0} & 79.7\deltapos{-9.8} && 23.1\deltapos{-3.1} & 78.0\deltapos{-1.4} && 78.8\deltapos{-5.2} & 101.1\deltaneg{+0.7} && 40.6\deltapos{-1.1} & 83.7\deltaneg{+1.9} \\
    \midrule
    $\beta = \beta_{\text{KL}}$ && 14.9\deltapos{-7.0} & 49.1\deltapos{-13.2} && 27.0\deltapos{-4.2} & 75.5\deltapos{-1.7} && 36.9\deltapos{-8.5} & 84.2\deltapos{-5.3} && 21.5\deltapos{-4.7} & 76.6\deltapos{-2.8} && 82.9\deltapos{-1.1} & 100.3\deltapos{-0.1} && 36.6\deltapos{-5.1} & 77.1\deltapos{-4.7} \\
    $\beta = \beta_{\text{H}}$  && 14.7\deltapos{-7.2} & 48.7\deltapos{-13.6} && 28.0\deltapos{-3.2} & 77.4\deltaneg{+0.2} && 36.7\deltapos{-8.7} & 84.1\deltapos{-5.4} && 21.5\deltapos{-4.7} & 76.6\deltapos{-2.8} && 82.9\deltapos{-1.1} & 100.1\deltapos{-0.3} && 36.8\deltapos{-4.9} & 77.4\deltapos{-4.4} \\
    $\beta = \beta_{\text{J}}$ && 21.2\deltapos{-0.7} & 60.0\deltapos{-2.3} && 26.3\deltapos{-4.9} & 74.0\deltapos{-3.2} && 36.8\deltapos{-8.6} & 84.3\deltapos{-5.2} && 21.5\deltapos{-4.7} & 76.5\deltapos{-2.9} && 82.8\deltapos{-1.2} & 100.2\deltapos{-0.2} && 37.7\deltapos{-4.0} & 79.0\deltapos{-2.8} \\
    \midrule
    \rowcolor{gray!20}
    \multicolumn{3}{l}{\textbf{Multiple analogy}} & && & && & && & && & && & \\
    \midrule
    $\beta=1$ && 17.2\deltapos{-4.7} & 53.9\deltapos{-8.4} && 22.1\deltapos{-9.1} & 67.4\deltapos{-9.8} && 29.7\deltapos{-15.7} & 74.6\deltapos{-14.9} && 19.5\deltapos{-6.7} & 74.0\deltapos{-5.4} && 79.8\deltapos{-4.2} & 100.9\deltaneg{+0.5} && 33.7\deltapos{-8.0} & 74.2\deltapos{-7.6} \\
    $\beta=1/N$  && 18.6\deltapos{-3.3} & 57.6\deltapos{-4.7} && 24.7\deltapos{-6.5} & 74.0\deltapos{-3.2} && 28.1\deltapos{-17.3} & 71.8\deltapos{-17.7} && 18.9\deltapos{-7.3} & 72.4\deltapos{-7.0} && 80.2\deltapos{-3.8} & 101.2\deltaneg{+0.8} && 34.1\deltapos{-7.6} & 75.4\deltapos{-6.4} \\
    \midrule
    $\beta = \beta_{\text{KL}}$ && 17.8\deltapos{-4.1} & 55.8\deltapos{-6.5} && 21.9\deltapos{-9.3} & 66.9\deltapos{-10.3} && 29.8\deltapos{-15.6} & 73.9\deltapos{-15.6} && 19.3\deltapos{-6.9} & 73.0\deltapos{-6.4} && 81.8\deltapos{-2.2} & 100.5\deltaneg{+0.1} && 34.1\deltapos{-7.6} & 74.0\deltapos{-7.8} \\
    $\beta = \beta_{\text{H}}$ && 18.4\deltapos{-3.5} & 57.1\deltapos{-5.2} && 22.2\deltapos{-9.0} & 67.9\deltapos{-9.3} && 28.2\deltapos{-17.2} & 71.2\deltapos{-18.3} && 18.9\deltapos{-7.3} & 72.4\deltapos{-7.0} && 80.2\deltapos{-3.8} & 100.9\deltaneg{+0.5} && 33.6\deltapos{-8.1} & 73.9\deltapos{-7.9} \\
    $\beta = \beta_{\text{J}}$ && 18.5\deltapos{-3.4} & 57.3\deltapos{-5.0} && 23.6\deltapos{-7.6} & 71.2\deltapos{-6.0} && 28.4\deltapos{-17.0} & 72.0\deltapos{-17.5} && 18.8\deltapos{-7.4} & 72.3\deltapos{-7.1} && 80.5\deltapos{-3.5} & 100.9\deltaneg{+0.5} && 34.0\deltapos{-7.7} & 74.7\deltapos{-7.1} \\
   
    \bottomrule
    \end{tabular}}
\end{table*}

\begin{table*}[t]
    \caption{Linear Probing results obtained with the best performing Linguistic Aware Single-Analogy and Multi-Analogy merged models, after fine-tuning only the final linear layer.}
    \centering
    \setlength{\tabcolsep}{.3em}
    \resizebox{\linewidth}{!}{%
    \footnotesize
    \begin{tabular}{l l ll l ll l ll l ll l ll l ll}
    \toprule
    && \multicolumn{2}{c}{\makecell{\textbf{IAM (en)}}} 
    && \multicolumn{2}{c}{\makecell{\textbf{Rimes (fr)}}}
    && \multicolumn{2}{c}{\makecell{\textbf{LAM (it)}}} 
    && \multicolumn{2}{c}{\makecell{\textbf{Rodrigo (es)}}} 
    && \multicolumn{2}{c}{\makecell{\textbf{$\text{ICFHR}_{2016}\text{ (de)}$}}} 
    && \multicolumn{2}{c}{\makecell{\textbf{AVG.}}} \\
    \cmidrule{3-4} \cmidrule{6-7} \cmidrule{9-10} \cmidrule{12-13} \cmidrule{15-16} \cmidrule{18-19}
    && \textbf{CER} $\downarrow$ & \textbf{WER} $\downarrow$ && \textbf{CER} $\downarrow$ & \textbf{WER} $\downarrow$ && \textbf{CER} $\downarrow$ & \textbf{WER} $\downarrow$ && \textbf{CER} $\downarrow$ & \textbf{WER} $\downarrow$ && \textbf{CER} $\downarrow$ & \textbf{WER} $\downarrow$ && \textbf{CER} $\downarrow$ & \textbf{WER} $\downarrow$ \\
    \midrule
    \multicolumn{19}{c}{\textbf{TrOCR (S)}}\\
    \midrule
    In-domain && 7.7 & 21.8 && 4.6 & 12.5 && 4.1 & 13.2 && 2.1 & 10.7 && 6.3 & 25.2 && 5.0 & 16.7 \\
    Baseline $\theta_T^{syn}$ && 13.3\deltazero & 33.1\deltazero && 13.1\deltazero & 29.9\deltazero && 37.5\deltazero & 66.4\deltazero && 18.0\deltazero & 48.5\deltazero && 70.0\deltazero & 99.0\deltazero && 30.4\deltazero & 55.4\deltazero \\
    \midrule
    \rowcolor{gray!20}
    \multicolumn{3}{l}{\textbf{Single analogy}} &&& & && & && & && & && & \\
    $\beta=\beta^{*}$  && 12.9\deltapos{-0.4} & 31.9\deltapos{-1.2} && 12.7\deltapos{-0.4} & 29.0\deltapos{-0.9} && 23.7\deltapos{-13.8} & 50.0\deltapos{-16.4} && 11.8\deltapos{-6.2} & 35.4\deltapos{-13.1} && 66.9\deltapos{-3.1} & 96.5\deltapos{-2.5} && 25.6\deltapos{-4.8} & 48.6\deltapos{-6.8} \\
    \midrule
    \rowcolor{gray!20}
    \multicolumn{3}{l}{\textbf{Multi analogy}} &&& & && & && & && & && & \\
    $\beta=\beta^{*}$ && 12.0\deltapos{-1.3} & 30.9\deltapos{-2.2} && 12.0\deltapos{-1.1} & 26.8\deltapos{-3.1} && 17.5\deltapos{-20.0} & 41.3\deltapos{-25.1} && 11.9\deltapos{-6.1} & 35.0\deltapos{-13.5} && 65.3\deltapos{-4.7} & 95.9\deltapos{-3.1} && 23.7\deltapos{-6.7} & 46.0\deltapos{-9.4} \\
    \midrule
    \multicolumn{19}{c}{\textbf{TrOCR (B)}}\\
    \midrule
    In-domain && 6.3 & 17.7 && 3.8 & 9.6 && 3.5 & 11.0 && 2.2 & 10.7 && 6.4 & 24.8 && 4.4 & 14.8 \\
    Baseline $\theta_T^{syn}$ && 12.3\deltazero & 30.1\deltazero && 9.5\deltazero & 21.9\deltazero && 26.0\deltazero & 52.8\deltazero && 15.3\deltazero & 41.2\deltazero && 63.9\deltazero & 94.7\deltazero && 25.4\deltazero & 48.1\deltazero \\
    \midrule
    \rowcolor{gray!20}
    \multicolumn{3}{l}{\textbf{Single analogy}} &&& & && & && & && & && & \\
    $\beta=\beta^{*}$  && 11.1\deltapos{-1.2} & 28.0\deltapos{-2.1} && 8.7\deltapos{-0.8} & 19.9\deltapos{-2.0} && 18.8\deltapos{-7.2} & 41.6\deltapos{-11.2} && 8.8\deltapos{-6.5} & 27.0\deltapos{-14.2} && 58.4\deltapos{-5.5} & 89.7\deltapos{-5.0} && 21.2\deltapos{-4.2} & 41.2\deltapos{-6.9} \\
    \midrule
    \rowcolor{gray!20}
    \multicolumn{3}{l}{\textbf{Multi analogy}} &&& & && & && & && & && & \\
    $\beta=\beta^{*}$ && 10.3\deltapos{-2.0} & 27.0\deltapos{-3.1} && 7.5\deltapos{-2.0} & 17.5\deltapos{-4.4} && 13.9\deltapos{-12.1} & 33.8\deltapos{-19.0} && 8.0\deltapos{-7.3} & 25.0\deltapos{-16.2} && 49.5\deltapos{-14.4} & 82.7\deltapos{-12.0} && 17.8\deltapos{-7.6} & 37.2\deltapos{-10.9} \\
    \midrule
    \multicolumn{19}{c}{\textbf{TrOCR (L)}}\\
    \midrule
    In-domain && 6.7 & 19.3 && 3.9 & 9.6 && 3.7 & 11.9 && 1.8 & 9.0 && 5.2 & 20.5 && 4.3 & 14.1 \\
    Baseline $\theta_T^{syn}$ && 11.6\deltazero & 29.5\deltazero && 9.1\deltazero & 21.0\deltazero && 28.6\deltazero & 55.4\deltazero && 16.2\deltazero & 44.2\deltazero && 61.6\deltazero & 92.8\deltazero && 25.4\deltazero & 48.6\deltazero \\
    \midrule
    \rowcolor{gray!20}
    \multicolumn{3}{l}{\textbf{Single analogy}} &&& & && & && & && & && & \\
    $\beta=\beta^{*}$  && 11.4\deltapos{-0.2} & 30.1\deltaneg{+0.6} && 8.3\deltapos{-0.8} & 18.4\deltapos{-2.6} && 11.8\deltapos{-16.8} & 30.2\deltapos{-25.2} && 12.7\deltapos{-3.5} & 35.8\deltapos{-8.4} && 52.5\deltapos{-9.1} & 84.5\deltapos{-8.3} && 19.3\deltapos{-6.1} & 39.8\deltapos{-8.8} \\
    \midrule
    \rowcolor{gray!20}
    \multicolumn{3}{l}{\textbf{Multi analogy}} &&& & && & && & && & && & \\
    $\beta=\beta^{*}$ && 11.4\deltapos{-0.2} & 29.6\deltaneg{+0.1} && 7.5\deltapos{-1.6} & 17.7\deltapos{-3.3} && 11.4\deltapos{-17.2} & 29.6\deltapos{-25.8} && 12.3\deltapos{-3.9} & 34.1\deltapos{-10.1} && 52.5\deltapos{-9.1} & 84.4\deltapos{-8.4} && 19.0\deltapos{-6.4} & 39.1\deltapos{-9.5} \\
    \midrule
    \multicolumn{19}{c}{\textbf{CRNN}}\\
    \midrule
    In-domain && 6.2 & 22.0 && 4.1 & 13.8 && 3.5 & 12.5 && 1.0 & 5.5 && 6.3 & 26.1 && 4.2 & 16.0 \\
    Baseline $\theta_T^{syn}$ && 15.9\deltazero & 46.0\deltazero && 17.6\deltazero & 51.5\deltazero && 31.1\deltazero & 72.0\deltazero && 18.5\deltazero & 65.7\deltazero && 72.4\deltazero & 100.0\deltazero && 31.1\deltazero & 67.0\deltazero \\
    \midrule
    \rowcolor{gray!20}
    \multicolumn{3}{l}{\textbf{Single analogy}} &&& & && & && & && & && & \\
    $\beta=\beta^{*}$  && 14.4\deltapos{-1.5} & 42.9\deltapos{-3.1} && 18.2\deltaneg{+0.6} & 52.6\deltaneg{+1.1} && 28.2\deltapos{-2.9} & 67.5\deltapos{-4.5} && 12.0\deltapos{-6.5} & 51.9\deltapos{-13.8} && 71.9\deltapos{-0.5} & 100.1\deltaneg{+0.1} && 28.9\deltapos{-2.2} & 63.0\deltapos{-4.0} \\
    \midrule
    \rowcolor{gray!20}
    \multicolumn{3}{l}{\textbf{Multi analogy}} &&& & && & && & && & && & \\
    $\beta=\beta^{*}$ && 12.6\deltapos{-3.3} & 39.1\deltapos{-6.9} && 15.4\deltapos{-2.2} & 47.0\deltapos{-4.5} && 24.4\deltapos{-6.7} & 62.0\deltapos{-10.0} && 12.8\deltapos{-5.7} & 54.3\deltapos{-11.4} && 64.8\deltapos{-7.6} & 100.2\deltaneg{+0.2} && 26.0\deltapos{-5.1} & 60.5\deltapos{-6.5} \\
    \midrule
    \multicolumn{19}{c}{\textbf{VAN}}\\
    \midrule
    In-domain && 6.7 & 25.5 && 5.5 & 20.4 && 3.7 & 13.5 && 1.7 & 11.2 && 9.4 & 38.5 && 5.4 & 21.8 \\
    Baseline $\theta_T^{syn}$ && 14.5\deltazero & 45.8\deltazero && 15.7\deltazero & 51.0\deltazero && 22.8\deltazero & 63.7\deltazero && 13.7\deltazero & 60.5\deltazero && 60.9\deltazero & 102.1\deltazero && 25.5\deltazero & 64.6\deltazero \\
    \midrule
    \rowcolor{gray!20}
    \multicolumn{3}{l}{\textbf{Single analogy}} &&& & && & && & && & && & \\
    $\beta=\beta^{*}$  && 14.0\deltapos{-0.5} & 45.1\deltapos{-0.7} && 16.5\deltaneg{+0.8} & 53.3\deltaneg{+2.3} && 18.9\deltapos{-3.9} & 55.9\deltapos{-7.8} && 9.3\deltapos{-4.4} & 47.3\deltapos{-13.2} && 52.3\deltapos{-8.6} & 100.4\deltapos{-1.7} && 22.2\deltapos{-3.3} & 60.4\deltapos{-4.2} \\
    \midrule
    \rowcolor{gray!20}
    \multicolumn{3}{l}{\textbf{Multi analogy}} &&& & && & && & && & && & \\
    $\beta=\beta^{*}$ && 13.4\deltapos{-1.1} & 43.9\deltapos{-1.9} && 18.1\deltaneg{+2.4} & 55.2\deltaneg{+4.2} && 15.0\deltapos{-7.8} & 47.2\deltapos{-16.5} && 9.2\deltapos{-4.5} & 46.3\deltapos{-14.2} && 50.0\deltapos{-10.9} & 99.2\deltapos{-2.9} && 21.1\deltapos{-4.4} & 58.4\deltapos{-6.2} \\
    \midrule
    \multicolumn{19}{c}{\textbf{HTR-VIT}}\\
    \midrule
    In-domain && 7.9 & 29.6 && 5.5 & 20.4 && 3.6 & 13.3 && 2.0 & 12.9 && 7.5 & 33.0 && 5.3 & 21.8 \\
    Baseline $\theta_T^{syn}$ && 15.9\deltazero & 50.7\deltazero && 17.5\deltazero & 56.7\deltazero && 20.6\deltazero & 60.6\deltazero && 9.6\deltazero & 50.5\deltazero && 41.4\deltazero & 94.6\deltazero && 21.0\deltazero & 62.6\deltazero \\
    \midrule
    \rowcolor{gray!20}
    \multicolumn{3}{l}{\textbf{Single analogy}} &&& & && & && & && & && & \\
    $\beta=\beta^{*}$  && 14.4\deltapos{-1.5} & 47.2\deltapos{-3.5} && 15.8\deltapos{-1.7} & 52.6\deltapos{-4.1} && 16.9\deltapos{-3.7} & 52.6\deltapos{-8.0} && 6.4\deltapos{-3.2} & 38.3\deltapos{-12.2} && 37.0\deltapos{-4.4} & 92.0\deltapos{-2.6} && 18.1\deltapos{-2.9} & 56.5\deltapos{-6.1} \\
    \midrule
    \rowcolor{gray!20}
    \multicolumn{3}{l}{\textbf{Multi analogy}} &&& & && & && & && & && & \\
    $\beta=\beta^{*}$ && 11.9\deltapos{-4.0} & 40.7\deltapos{-10.0} && 13.3\deltapos{-4.2} & 45.3\deltapos{-11.4} && 13.9\deltapos{-6.7} & 43.9\deltapos{-16.7} && 5.2\deltapos{-4.4} & 31.8\deltapos{-18.7} && 31.6\deltapos{-9.8} & 88.0\deltapos{-6.6} && 15.2\deltapos{-5.8} & 49.9\deltapos{-12.7} \\
    \bottomrule
    \end{tabular}}    
\label{tab:LP_all}
\end{table*}

\end{document}